\documentclass[pdflatex,sn-mathphys-num]{sn-jnl}

\usepackage{graphicx}%
\usepackage{multirow}%
\usepackage{multicol}
\usepackage{tabularx}
\usepackage{amsmath,amssymb,amsfonts}%
\usepackage{amsthm}%
\usepackage{mathrsfs}%
\usepackage[title]{appendix}%
\usepackage{xcolor}%
\usepackage{textcomp}%
\usepackage{manyfoot}%
\usepackage{booktabs}%
\usepackage{algorithm}%
\usepackage{algorithmicx}%
\usepackage{algpseudocode}%
\usepackage{listings}%
\usepackage{caption}   

\theoremstyle{thmstyleone}%
\theoremstyle{thmstyletwo}%

\theoremstyle{thmstylethree}%

\begin{document}

\title[Article Title]{Advancing Histopathology-Based Breast Cancer Diagnosis: Insights into Multi-Modality and Explainability}


\author*[1]{\fnm{Faseela} \sur{Abdullakutty}}\email{faseela.abdullakutty@qu.edu.qa}

\author*[1]{\fnm{Younes} \sur{Akbari}}\email{akbari\_younes@yahoo.com}

\author*[1]{\fnm{Somaya} \sur{  Al-Maadeed}}\email{s\_alali@qu.edu.qa}
\author[2]{\fnm{Ahmed } \sur{   Bouridane}}\email{abouridane@sharjah.ac.ae}
\author[2]{\fnm{Rifat} \sur{  Hamoudi}}\email{rhamoudi@sharjah.ac.ae}
\affil*[1]{ \orgname{Qatar University},   \country{Qatar}}

\affil[2]{ \orgname{University of Sharjah},  \country{United Arab Emirates}}



\abstract{ It is imperative that breast cancer is detected precisely and timely to improve patient outcomes. Diagnostic methodologies have traditionally relied on unimodal approaches; however, medical data analytics is integrating diverse data sources beyond conventional imaging. Using multi-modal techniques, integrating both image and non-image data, marks a transformative advancement in breast cancer diagnosis. The purpose of this review is to explore the burgeoning field of multimodal techniques, particularly the fusion of histopathology images with non-image data. Further, Explainable AI (XAI) will be used to elucidate the decision-making processes of complex algorithms, emphasizing the necessity of explainability in diagnostic processes. This review utilizes multi-modal data and emphasizes explainability to enhance diagnostic accuracy, clinician confidence, and patient engagement, ultimately fostering more personalized treatment strategies for breast cancer, while also identifying research gaps in multi-modality and explainability, guiding future studies, and contributing to the strategic direction of the field.}




\maketitle

\section{Introduction}\label{intro}

In the realm of breast cancer diagnosis, the convergence of multi-modal techniques, amalgamating both image and non-image data \cite{sun2023scoping}, heralds a transformative approach with profound implications for disease detection and characterization. As a leading cause of mortality among women globally, the precise and timely diagnosis of breast cancer remains imperative for optimizing patient outcomes. While traditional diagnostic methodologies \cite{krithiga2021breast} have historically relied heavily on uni-modal approaches, the evolving landscape of medical data analytics underscores the significance of integrating diverse data sources beyond conventional imaging modalities \cite{tafavvoghi2024publicly}.

\begin{figure}[H]

  \centering
  
  \includegraphics[width=\textwidth]{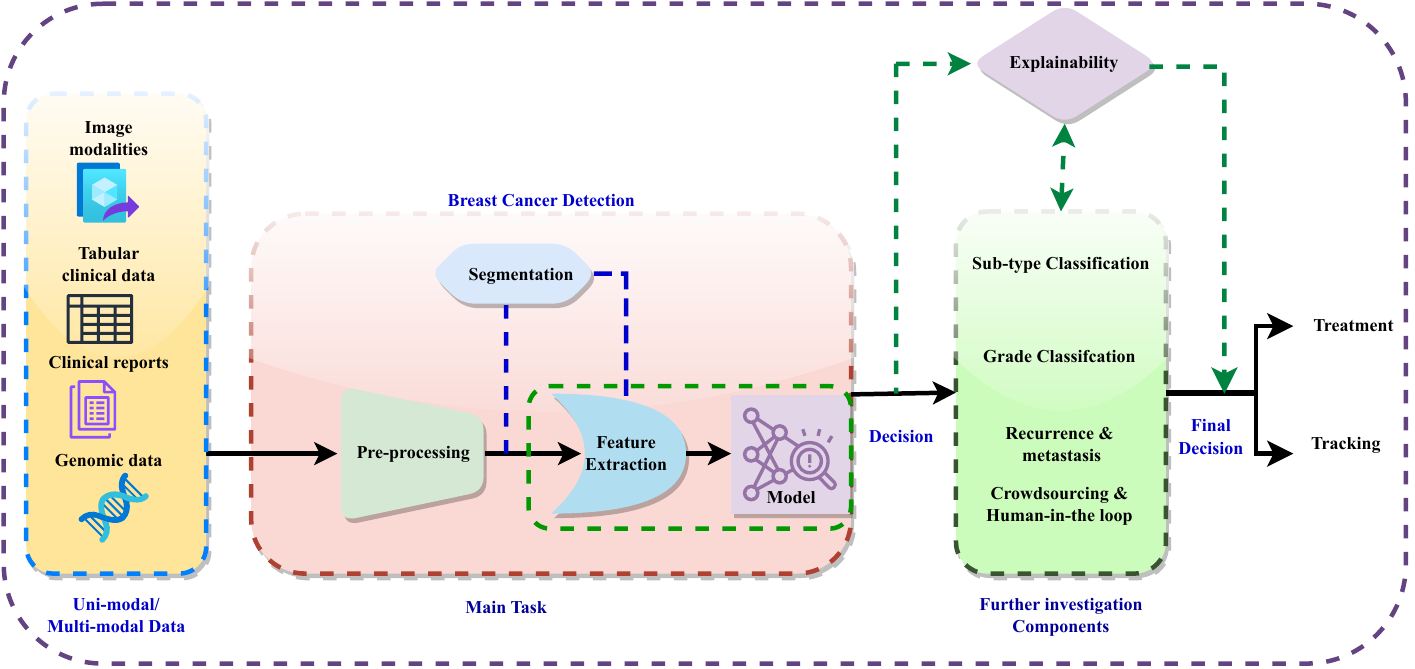}
 
  \caption{A generic representation of breast cancer diagnosis}
  \label{fig:gen_bcd}

\end{figure}

\begin{figure}[H]

  \centering
  
  \includegraphics[width=\textwidth]{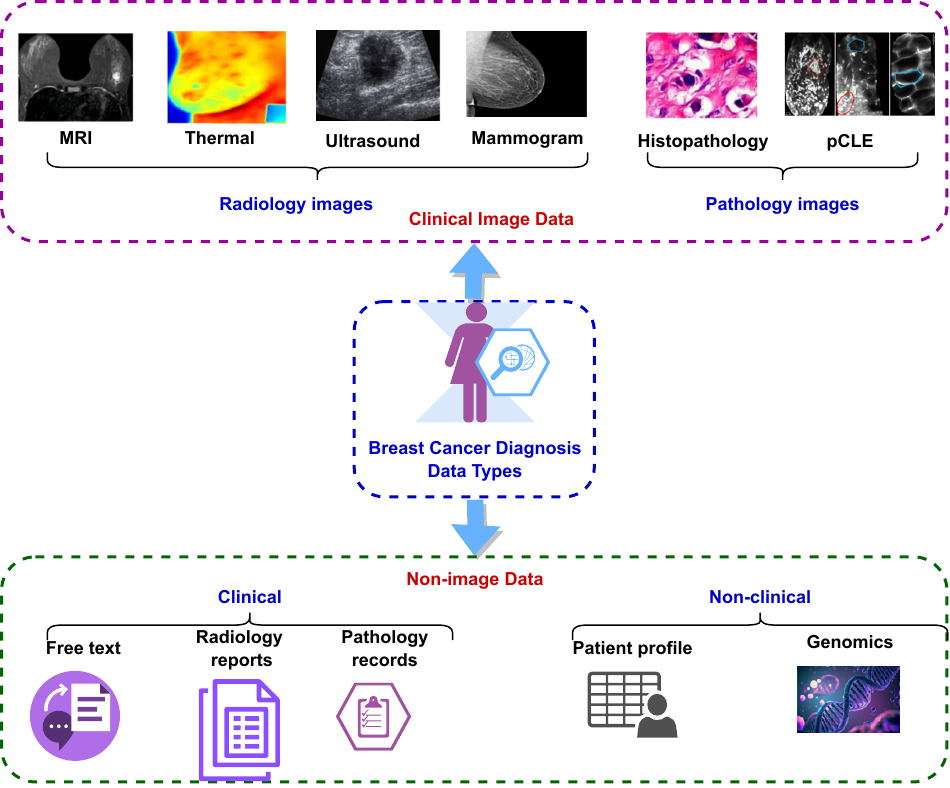}
 
  \caption{Types of breast cancer diagnosis data }
  \label{fig:mdata_type}
  \end{figure}

Figure \ref{fig:gen_bcd} illustrates a generic model for breast cancer diagnosis within the Computer-Aided Detection (CAD) framework. As depicted in Figure \ref{fig:mdata_type}, breast cancer detection can be performed using various data types, employing either unimodal or multimodal approaches. The process initiates with data pre-processing, followed by feature extraction. To enhance the learning of feature representations from image data, segmentation may be conducted prior to feature extraction. Subsequently, the detection model is applied to generate a diagnosis from the processed data. Based on this diagnosis, further analyses are performed, including sub-type classification, grade classification, recurrence and metastasis prediction, as well as the incorporation of crowdsourcing and human-in-the-loop methodologies. These steps culminate in a final decision that informs subsequent treatment and monitoring strategies.

This review endeavours to illuminate the burgeoning field of multi-modal techniques in breast cancer diagnosis context, placing particular emphasis on the fusion of heterogeneous data streams encompassing both image and non-image modalities. Beyond the confines of traditional imaging modalities, such as mammography, magnetic resonance imaging (MRI), ultrasound, and positron emission tomography (PET), multi-modal approaches leverage a plethora of non-image data types including genetic markers, proteomic profiles, clinical parameters, and patient demographics \cite{abo2024advances,hussain2024breast}. By harnessing the complementary insights gleaned from these diverse data modalities, multi-modal techniques offer a multifaceted understanding of breast cancer biology and pathology, transcending the limitations of uni-modal approaches.

Furthermore, alongside the integration of multi-modal data, the exigency for explainability in breast cancer diagnosis emerges as a pivotal consideration. As machine learning and artificial intelligence algorithms increasingly permeate diagnostic workflows, the interpretability and transparency of decision-making processes assume paramount importance. Explainable AI (XAI) techniques endeavor to demystify the opaque nature of complex algorithms, elucidating the rationale behind diagnostic decisions and enhancing the interpretability of diagnostic outcomes \cite{hussain2024breast}. In the context of breast cancer diagnosis, explainability not only engenders clinician confidence in decision support systems but also fosters patient understanding and engagement, empowering informed decision-making and facilitating personalized treatment strategies.

\begin{table}[ht]
\caption{Latest reviews on breast cancer diagnosis in various contexts. }
\label{tab:reviews}

\begin{tabular}{lclccc}
\hline
\textbf{Author}                                                   & \textbf{Year} & \textbf{Main Discussion}                                                                                                       & \textbf{Datasets} & \textbf{\begin{tabular}[c]{@{}l@{}}Multi \\ modality \end{tabular}} & \textbf{XAI} \\ \hline
Abo-El-Rejalet al.\cite{abo2024advances}        & 2024          & Segmentation                                                                                                                   & $\times $              & $\times $                    & $\times $                   \\
Brodhead et al.\cite{brodhead2024multimodality} & 2024          & Imaging characteristics                                                                                                        & $\times $              & $\times $                    & $\times $                    \\
Hussain et al.\cite{hussain2024breast}          & 2024          & Breast cancer risk prediction                                                                                                  & $\times $             & $\checkmark$                       & $\checkmark$                       \\
Luo et al.\cite{luo2024deep}                     & 2024          & Breast Cancer Imaging                                                                                                          & $\times $              & $\checkmark$                       & $\times $                    \\
Rautela et al.\cite{rautela2024comprehensive}   & 2024          & \begin{tabular}[c]{@{}l@{}}Computational techniques\\ for breast cancer\end{tabular}                                           & $\times $             & $\checkmark$                       & $\times $                    \\
Singh et al.\cite{singh2024technical}           & 2024          & \begin{tabular}[c]{@{}l@{}}Breast Cancer Screening and\\ Detection using \\Artificial Intelligence\\  and Radiomics\end{tabular} & $\times $              & $\checkmark$                       & $\times $                    \\
Thakur et al .\cite{thakur2024systematic}        & 2024          & \begin{tabular}[c]{@{}l@{}}Identification and of breast\\ cancer through medical \\ image modalities\end{tabular}                 & $\checkmark$                & $\checkmark$                       & $\times $                    \\ \hline
\end{tabular}
\end{table}

Table \ref{tab:reviews} presents recent reviews on breast cancer diagnosis across various contexts. However, these reviews often overlook multi-modality and explainability, treating them as future research directions rather than discussing existing methods. Additionally, there is a lack of focus on histopathology and frameworks that combine histopathology with non-image data for breast cancer detection. In light of these observations, this review addresses multi-modal datasets, including histopathology and other non-image data, explores multi-modal techniques utilizing these datasets, and examines explainable multi-modal methods in histopathology-based breast cancer diagnosis.

The major contributions of this article are:

\begin{itemize}
    \item A  detailed investigation of multi-modal datasets, including those that incorporate histopathology and non-image data, which are frequently overlooked in existing literature.
    \item A discussion on multi-modal techniques that utilize the aforementioned datasets, offering insights into their application and effectiveness in breast cancer diagnosis.
    \item An investigation of explainable multi-modal methods specifically within the context of histopathology-based breast cancer diagnosis, addressing a critical gap in current research.
    \item Identification research gaps in multi-modality and explainability, identifying key areas for future study and contributing to the strategic direction of the field.
   
\end{itemize}






\section{Breast Cancer Diagnosis: An Overview}
Breast cancer is among the most prevalent malignancies impacting women globally, presenting substantial challenges for public health and individual well-being. The importance of early diagnosis cannot be overstated, as it is pivotal in enhancing treatment efficacy and elevating survival rates. This overview delves into the diverse tasks involved in breast cancer diagnosis. By critically examining the current methodologies and advancements in breast cancer detection \cite{nasser2023deep}, strategies for ensuring prompt, precise, and effective diagnostic processes can be elucidated, thereby fostering improved patient care and clinical outcomes.
\subsection{Tasks}

\begin{figure}[H]

  \centering
  
  \includegraphics[width=.85\textwidth,height=8.5cm]{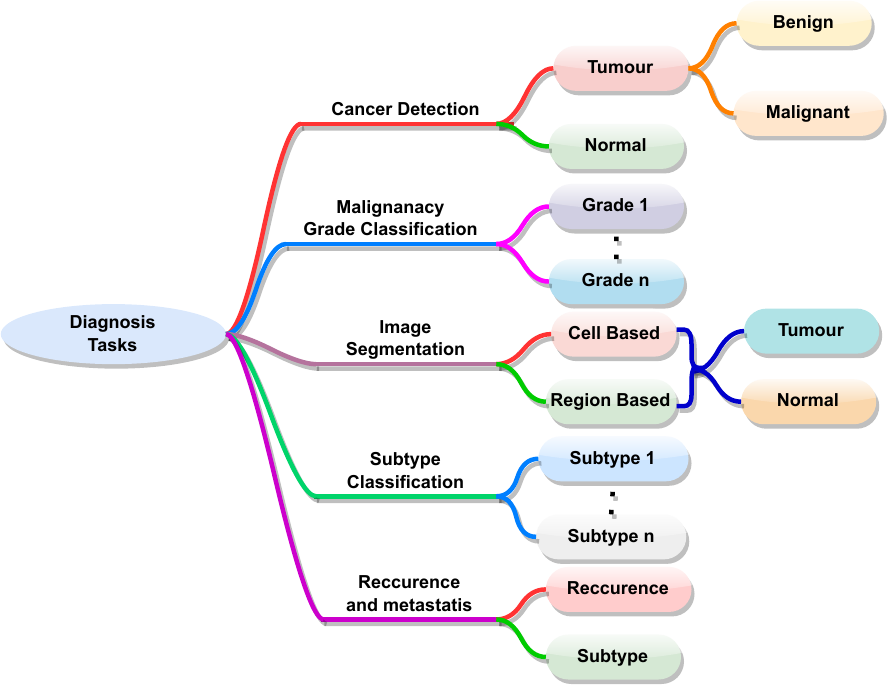}
 
  \caption{Breast Cancer diagnosis tasks }
  \label{fig:tasks}

\end{figure}

Breast cancer diagnosis \cite{obeagu2024breast} involves several tasks that can utilize image and non-image data as shown in Figure. \ref{fig:tasks}. 
Breast cancer detection entails pinpointing signs of cancer within breast imaging data, such as mammograms or ultrasound scans. By harnessing machine learning algorithms, it's possible to analyze these images to identify potentially suspicious areas or anomalies that could suggest the presence of tumours. These advanced techniques \cite{rai2024cancer} offer a more efficient and potentially more accurate method for detecting early signs of breast cancer, providing valuable insights for healthcare professionals in their diagnostic process.

Malignancy classification \cite{liu2023classifier} is the process of determining whether detected abnormalities are malignant, indicating cancer, or benign, meaning they are non-cancerous. This step is vital for guiding the subsequent treatment plan. Machine learning models can assist in this classification by analyzing features derived from imaging data, including characteristics like shape, texture, and intensity. By training these models on large datasets, they can provide predictions on the probability that an abnormality is cancerous, aiding healthcare professionals in making informed decisions regarding patient care \cite{kumaraswamy2023invasive}.

Subtype classification is a crucial process in understanding breast cancer, as it encompasses a spectrum of diseases, each with unique traits and outcomes \cite{huang2024classifying}. This step involves dividing breast cancer cases into specific subtypes like hormone receptor-positive, HER2-positive, or triple-negative breast cancer, which are known to have varying responses to treatments and differing prognoses. By categorizing cases into these subtypes, medical professionals can tailor treatment plans more effectively \cite{choi2023mobrca}. Machine learning models play a role in this by analyzing genomic data, gene expression profiles, and clinical information to predict the subtype, facilitating personalized and targeted therapeutic approaches.

\begin{table}[!ht]
\caption{Recent research in breast cancer diagnosis including different tasks }
\label{tab:task_gen}
\begin{tabular}{llll}
\hline
\textbf{Method}                                                                                                                  & \textbf{Dataset}                                                                                                                                                         & \textbf{Modality}                                                            & \textbf{Task}                                                       \\ \hline
\begin{tabular}[c]{@{}l@{}}Classifier-combined\\ method \end{tabular}\cite{liu2023classifier}                                                             & Proprietary                                                                                                                                                      & MRI                                                                          &\begin{tabular}[c]{@{}l@{}} Grade \\ Classification  \end{tabular}                                              \\
DeepBreastCancerNet \cite{raza2023deepbreastcancernet}                                                          & \begin{tabular}[c]{@{}l@{}}BUSI \cite{al2020dataset},\\ \begin{tabular}[c]{@{}l@{}}Ultrasound \\ Image dataset \end{tabular}\cite{paulo2017breast}\end{tabular}                   & Ultrasound                                                                   & Detection                                                           \\
DSCCN \cite{huang2024classifying}                                                                               & TCGA   \cite{tcga2023}                                                                                                                                                                  & multi-omics                                                                  & \begin{tabular}[c]{@{}l@{}} Sub-type \\ Classification  \end{tabular}                                            \\
EMDCOC \cite{parshionikar2024enhanced}                                                                          & \begin{tabular}[c]{@{}l@{}}BreakHis  \cite{spanhol2015dataset}   \\ \begin{tabular}[c]{@{}l@{}}IR Thermal\\ Images\end{tabular} \cite{zuluaga2021cnn}\end{tabular}                                                              & \begin{tabular}[c]{@{}l@{}}Histopathology,\\  IR thermal images\end{tabular} & Detection                                                           \\
Ensemble CNN \cite{kumaraswamy2023invasive}                                                                     & Databiox \cite{databiox_datasets}                                                                                                                      & Histopathology                                                               & \begin{tabular}[c]{@{}l@{}} Grade \\ Classification  \end{tabular}                                              \\
\begin{tabular}[c]{@{}l@{}}histogram K-means \\
segmentation \end{tabular} \cite{sahu2023cnn}                                                               & BreakHis \cite{spanhol2015dataset}                                                                                                                      & Histopathology                                                               & Segmentation                                                        \\
Hybrid CNN \cite{sahu2023high}                                                                                  & \begin{tabular}[c]{@{}l@{}}Mini-DDSM \cite{lekamlage2020mini}, \\ BUSI \cite{al2020dataset}\end{tabular}                               & \begin{tabular}[c]{@{}l@{}}Mammogram, \\ Ultrasound images\end{tabular}      & Detection                                                           \\
Hybrid CNN-LSTM \cite{srikantamurthy2023classification}                                                         & BreakHis     \cite{spanhol2015dataset}                                                                                                                                                                & Histopathology                                                               & \begin{tabular}[c]{@{}l@{}} Grade \\ Classification  \end{tabular}                                            \\
KAMnet \cite{guo2024multimodal}                                                                                 & Proprietary                                                                                                                                                     & Ultrasound                                                                   & Detection                                                           \\
moBRCA‑net \cite{choi2023mobrca}                                                                                & TCGA   \cite{tcga2023}                                                                                                                                                                  & Multi-omics,                                                                 & \begin{tabular}[c]{@{}l@{}} Sub-type \\ Classification  \end{tabular}                                            \\
Multi-modal fusion \cite{liu2024multi}                                                                           & TCGA     \cite{tcga2023}                                                                                                                                                                & WSI, Gene Expression                                                         & Detection                                                           \\
\begin{tabular}[c]{@{}l@{}}optimized LSTM with \\ U-net segmentation\cite{sivamurugan2023applying}\end{tabular} & MIAS \cite{kendall2013automatic}                                                                                                                        & Mammogram                                                                    & Segmentation                                                        \\
\begin{tabular}[c]{@{}l@{}}Prediction model for \\ distant metastasis\cite{murata2023prediction}\end{tabular}   & Proprietary                                                                                                                                                      & Clinical Data                                                                & \begin{tabular}[c]{@{}l@{}}Reccurence\\ and metastatis\end{tabular} \\
recurrence prediction \cite{hussein2024framework}                                                               & WPBC                                                                                                                                                                     & Clinical Data                                                                & \begin{tabular}[c]{@{}l@{}}Recurrence\\ and metastasis\end{tabular} \\
Semantic Segmentation \cite{ahmed2023images}                                                                    & \begin{tabular}[c]{@{}l@{}}CBIS-DDSM \cite{lee2017curated}, \\ MIAS \cite{kendall2013automatic}\end{tabular}                           & Mammogram                                                                    & Segmentation                                                        \\
Unet3+  \cite{alam2023improving}                                                                                & Proprietary                                                                                                                                                    & Ultrasound                                                                   & Segmentation                                                        \\
Yolo‑Based Model \cite{prinzi2024yolo}                                                                          & \begin{tabular}[c]{@{}l@{}}CBIS-DDSM \cite{lee2017curated},\\ Inbreast \cite{moreira2012inbreast},\\  Proprietary \end{tabular} & Mammogram                                                                    & Detection                                                           \\ \bottomrule
\end{tabular}
\end{table}

Image segmentation \cite{rai2024cancer} involves dividing an image into cell segmentation and distinct segments or regions of interest. Within the realm of breast cancer diagnosis, segmentation helps to demarcate the boundaries of tumours or suspicious lesions in breast imaging data \cite{guo2024multi}. This process is critical for precisely measuring tumour size and shape, and it lays the groundwork for further analyses, including tumour volume estimation or extracting quantitative features. Machine learning algorithms, especially deep learning models like convolutional neural networks (CNNs), have demonstrated strong capabilities in automatically segmenting breast lesions from medical images \cite{rajoub2024segmentation}, offering a powerful tool to enhance the accuracy and efficiency of breast cancer diagnosis.

Predicting cancer recurrence and metastasis \cite{soliman2024artificial} is a crucial aspect of breast cancer management, extending beyond initial diagnosis and treatment. This task involves assessing the risk of the cancer returning or spreading to other parts of the body. Machine learning models can combine multiple types of data—such as imaging, genomic information, clinical variables (like patient demographics and medical history), and treatment records—to estimate the likelihood of recurrence or metastasis \cite{hussein2024framework}. These predictions are valuable for clinicians, allowing them to customize follow-up care and create personalized treatment plans for breast cancer patients, ultimately enhancing patient outcomes and reducing the risk of adverse events. It should be noted that the tasks should be combined and integrated to have an accurate system. For example, cancer detection for subtype classification should use the tasks of cancer segmentation and grading tasks and this process can improve the task of subtype classification \cite{gallagher2024artificial}.

Table. \ref{tab:task_gen} presents a summary of recent research advancements in breast cancer diagnosis across various tasks. A significant observation is the predominance of unimodal approaches in current methodologies. While some existing multimodal methods incorporate different types of imaging, such as ultrasound and mammography, the integration of image data with non-image data remains significantly underexplored. 
In particular, the fusion of histopathology images with non-image data, including textual and clinical information, represents a largely untapped area. The potential benefits of this integration are substantial. By combining histopathological imaging with comprehensive clinical and textual data, and leveraging advanced machine learning techniques, there is a strong potential to enhance the accuracy and efficiency of breast cancer diagnosis, prognosis, and treatment planning. This holistic approach could lead to significant advancements in personalized medicine and improved patient outcomes.

\subsection {Datasets}

The dataset used for breast cancer diagnosis encompasses both clinical image data and non-image data \cite{sweetlin2021review}, as illustrated in Figure \ref{fig:mdata_type}. The clinical image data comprise radiology and pathology images. Radiology images encompass modalities such as MRI, CT, thermal imaging, mammograms, and ultrasound, while pathology images include histopathology and pCLE \cite{tafavvoghi2024publicly}. The non-image data can be subdivided into clinical and non-clinical categories. Clinical data encompass radiology reports, pathology reports, including laboratory results, and narrative descriptions of patient status. Non-clinical data comprise patient profiles containing demographic information, patient history, age, other non-clinical details, and genomic data \cite{heiliger2023beyond}.

Additionally, non-image data are further classified into structured and unstructured categories. Radiology reports and narrative descriptions of patient status fall under unstructured data, while recorded pathology reports and patient profiles are considered structured data \cite{laokulrath2024invasive}.
Despite the abundance of both image and non-image datasets related to breast cancer detection, this paper focuses specifically on histopathology-based datasets, examining them in a multi-modal context. Table. \ref{tab:dataset}, lists the existing public datasets in breast cancer detection, based on histopathology. It is evident from the table that the number of multi-modal datasets is much less compared to the unimodal datasets. Also, the sample size is low in most of these datasets.

The landscape of breast cancer histopathology research is enriched by a diverse array of datasets, each offering unique features and clinical insights. Uni-modal datasets, such as BRACS \cite{brancati2022bracs}  and BreCaHAD \cite{aksac2019brecahad}, focus on a single type of data. The BRACS dataset provides 547 Whole-Slide Images (WSIs) and 4539 Regions Of Interest (ROIs), meticulously annotated by three board-certified pathologists. This dataset categorizes lesions into types such as Normal, Pathological Benign, Usual Ductal Hyperplasia, Flat Epithelial Atypia, Atypical Ductal Hyperplasia, Ductal Carcinoma in Situ, and Invasive Carcinoma. Similarly, the BreCaHAD dataset includes 162 histopathology images focusing on malignant cases, classified into mitosis, apoptosis, tumour nuclei, non-tumour nuclei, tubule, and non-tubule, thus facilitating comprehensive analyses and validation of diagnostic methods.


In contrast, multi-modal datasets integrate various data types to provide a more comprehensive view of breast cancer pathology. The TCGA-BRCA \cite{tcga2023} dataset, for instance, combines gene expression data, copy number variations (CNVs), and pathological images from 1098 breast cancer patients. This multi-dimensional approach allows for a deeper understanding of the molecular and histological characteristics of breast cancer. Similarly, the IMPRESS dataset includes Hematoxylin and Eosin (H\&E) and immunohistochemistry (IHC) stained WSIs from 126 patients, along with clinical data and biomarker annotations. The Post-NAT-BRCA38 dataset \cite{martel2019assessment} offers 96 WSIs along with detailed clinical information, including estrogen receptor (ER) status, progesterone receptor (PR) status, and human epidermal growth factor receptor 2 (HER2) status. These multi-modal datasets enable researchers to explore the interplay between genetic, molecular, and histological data, driving advancements in personalized breast cancer diagnosis and treatment.

\begin{table}[ht]
\caption{Multi-Modal Datasets public for Breast Cancer Diagnosis Featuring Histopathology Images}
\label{tab:dataset}
\begin{tabular}{lccl}
\hline
\textbf{Dataset}                                           & \textbf{Year} & \textbf{Size} & \textbf{Modalities}                                                                    \\ \hline
Post-NAT-BRCA \cite{martel2019assessment} & 2019          & 96            & WSI ,Clinical data                                                                     \\
CPTAC-BRCA \cite{cptac2020}               & 2020          & 642           & \begin{tabular}[c]{@{}l@{}}WSI, clinical, proteomic,\\ genomic data\end{tabular}       \\
Pathological EMR \cite{yan2021richer}     & 2021          &               & WSI, patient profile                                                                   \\
BCNB \cite{bcnb2022}                      & 2022          & 1,058         & Clinical Data                                                                          \\
IMPRESS \cite{huang2023artificial}        & 2023          & 126           & WSI ,Clinical data                                                                     \\
GTEx-Breast dataset \cite{gtex2023}       & 2023          & 894           & WSI, pathology notes                                                                   \\
TCGA-BRCA dataset \cite{tcga2023}         & 2023          & 1098          & \begin{tabular}[c]{@{}l@{}}WSI, \\ gene expression, CNV\end{tabular} \\ \hline
\end{tabular}
\end{table}

The IMPRESS dataset \cite{huang2023artificial} consists of 126 breast H\&E WSIs from 62 female patients with HER2-positive breast cancer and 64 female patients with triple-negative breast cancer, all of whom underwent neoadjuvant chemotherapy followed by surgical excision. It includes immunohistochemistry (IHC) stained WSIs of the same slides, along with corresponding scores. All slides were scanned using a Hamamatsu scanner at 20× magnification. The dataset also provides clinical data for both patient groups, including age, tumour size, and annotations for biomarkers such as PD-L1, CD-8, and CD-163.
The GTEx-Breast dataset \cite{gtex2023} is part of the Genotype-Tissue Expression (GTEx) project, which offers gene expression data across 44 human tissues. It includes 894 breast tissue histology images, comprising 306 WSIs of female breast tissue and 588 WSIs of male breast tissue, collected from the central subareolar region of the right breast at various centers in the United States. The images are accompanied by brief pathology notes and an annotation file with detailed sample information.

The CPTAC-BRCA dataset \cite{cptac2020}, from the Clinical Proteomic Tumor Analysis Consortium, includes 642 WSIs from 134 patients with breast invasive carcinoma, scanned at 20× magnification. The images are available in two resolutions: 0.25 $mu$m/pixel and 0.5 $mu$m/pixel. The dataset is accompanied by comprehensive clinical, proteomic, and genomic data.
The BCNB dataset \cite{bcnb2022}, or Early Breast Cancer Core-Needle Biopsy WSI Dataset, is the only publicly available collection of breast histopathological WSIs from Asia. It contains 1,058 WSIs from 1,058 breast cancer patients in China, scanned with an Iscan Coreo pathological scanner. Tumor regions in each image are annotated by two pathologists. The dataset also includes extensive clinical data such as patient age, tumour size, histological and molecular subtypes, number of lymph node metastases, and HER2, ER, and PR status.

\section{Exploring multi-modality}

Multi-modal techniques are increasingly significant in histopathology-based breast cancer detection due to their capability to enhance diagnostic accuracy, provide comprehensive insights, and improve patient outcomes. These techniques integrate various types of data and analytical methods, offering a more robust framework for detecting and characterizing breast cancer. Multimodal data fusion combines information from different modalities to improve decision-making processes, which is particularly useful in medical fields where single observations may yield diverse interpretations.
For instance, while isocitrate dehydrogenase (IDH1) mutation status and histological profiles individually contribute to understanding patient outcomes, their combined analysis has been crucial in revising the WHO classification of diffuse gliomas. Similarly, in breast cancer detection, artificial intelligence (AI) provides an automated and objective means to incorporate complementary information and clinical context from diverse datasets, thereby enhancing predictive accuracy.

\begin{figure}[H]

  \centering
  
  \includegraphics[width=\textwidth]{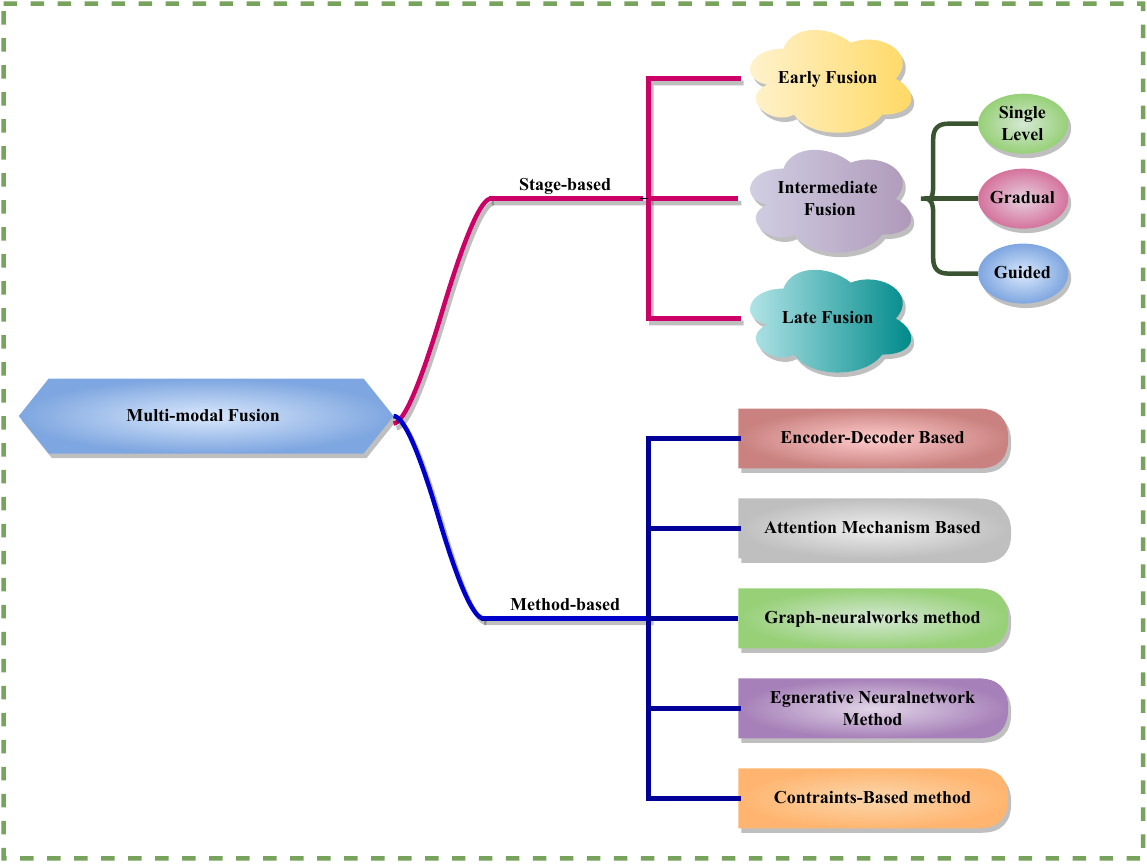}
 
  \caption{Multi-modal approaches}
  \label{fig:multi-modal}
  \end{figure}

As shown in Figure. \ref{fig:multi-modal}, the multi-modal fusion can be categorised as stage-based and method-based techniques. Stage-based 
 fusion strategies can be further categorized into early, late, and intermediate fusion approaches \cite{lipkova2022artificial}, each offering unique advantages in breast cancer detection. This approach is particularly beneficial when unimodal data are noisy or incomplete, as integrating redundant information from other modalities can improve the robustness and precision of predictions.

Multi-modal fusion approaches \cite{zhao2024deep}  include encoder-decoder methods, which combine feature extraction, fusion, and decision-making processes into a single model, making them efficient in tasks like video captioning and object detection. Attention mechanism methods use mechanisms like co-attention and cross-attention to enhance each modality with information from other modalities, allowing the model to fuse features and learn interdependencies among them. Graph Neural Network methods use GNN to capture long-range dependencies among different modalities, categorizing tasks into different classes based on data types. Generative Neural Network methods include models like VAE-based adversarial frameworks, which reduce distance differences between unimodal representations and are crucial for tasks like text-conditional image generation and image style transfer. Constrained-based methods involve innovative approaches like channel-exchanging-networks, which dynamically exchange channels in different modal sub-networks based on individual channel importance, but are limited to homogeneous data.

Multi-modal techniques can improve diagnostic accuracy by combining different data modalities such as histopathological images, molecular profiles, and clinical data. By integrating these data, they can differentiate between cancer subtypes, assess tumour heterogeneity, and predict potential treatment responses. They also contribute to better prognostication and personalized medicine by providing a holistic view of tumour biology and patient condition. 
Advanced imaging and computational tools, such as machine learning and artificial intelligence (AI), have revolutionized the analysis of histopathological data, automating the detection and classification of cancerous cells, extracting and analyzing complex patterns, and providing decision support to pathologists. In research, multi-modal techniques facilitate a deeper understanding of breast cancer mechanisms, leading to the discovery of new therapeutic targets and biomarkers. In clinical settings, these techniques enhance diagnostic workflows and support real-time decision-making, ultimately improving patient care.

Multi-modal techniques offer a more accurate, comprehensive, and personalized approach to breast cancer diagnosis and treatment, but they face challenges such as data standardization, computational resources management, and interdisciplinary collaboration. Future advancements in technology and computational methods are expected to address these challenges, making multimodal techniques more effective and widely adopted in clinical practice. However, challenges persist, such as the richness of feature representation \cite{yan2021richer} in images and the inadequacy of information fusion, which can lead to the loss of high-dimensional information and partially missing data in real-world scenarios. Each modality within multimodal data possesses distinct characteristics, adding to the complexity of heterogeneous data and further complicating multimodal fusion methods.

\subsection{Multi-Modal Breast Cancer Diagnosis: Combining Histopathology and Non-Image Modalities}
The integration of multi-modal approaches in breast cancer diagnosis, including histopathology and non-image modalities, improves diagnostic accuracy, provides a comprehensive understanding of the disease, improves personalized treatment planning, facilitates early detection and timely intervention, potentially improving patient outcomes, and promotes interdisciplinary collaboration among specialists. This approach reduces the likelihood of misdiagnosis, provides a more comprehensive understanding of tumor biology and patient health, and facilitates early detection and timely intervention, ultimately advancing clinical research. Table. \ref{tab:bc-mm} shows recent multi-modal research in breast cancer diagnosis.

Combining heterogeneous data without losing information from high-dimensional images has posed a significant challenge in data fusion. To address this issue, the authors of \cite{yan2021richer} proposed a multi-modal fusion technique that increased the dimensionality of structured data to align with the high-dimensional image features from histopathology WSI. They employed VGG-16 as the backbone model for image feature extraction and utilized a denoising autoencoder to enhance the dimensionality of the structured clinical data. The extracted features were then flattened and combined through concatenation before being input into fully connected layers for classification. This model successfully classified cases as benign or malignant, with its performance evaluated using the pathological electronic medical record (EMR) dataset.
PathLDM \cite{yellapragada2024pathldm}, is a text-conditioned Latent Diffusion Model designed to improve histopathology image generation by leveraging contextual information from pathology text reports. The model utilized Generalized Pathological Text (GPT) to distill and summarize complex text reports, thereby establishing an effective conditioning mechanism. The authors achieved a state-of-the-art Fréchet Inception Distance (FID) score of 7.64 for text-to-image generation on the TCGA-BRCA dataset, surpassing the closest text-conditioned competitor. The study compared PathLDM against alternative methods such as Moghadam et al., Medfusion, and Stable Diffusion, demonstrating its superiority in generating high-quality histopathology images conditioned on text.

Authors of \cite{ding2023improving} presented a method for improving mitosis detection in histopathology images using large vision-language models, combining image captioning and visual question-answering tasks with pre-trained models that integrate visual features and natural language. The inclusion of metadata, such as tumor and scanner types, into the question prompts significantly enhanced prediction accuracy. This method outperformed baseline models, including single-modality models and the vision-language model CLIP, demonstrating superior mitosis detection. Lu et al. (2023) \cite{lu2023visual} introduced the MI-Zero framework, which utilized contrastively aligned image and text models for zero-shot transfer on gigapixel histopathology whole slide images. Reformulating zero-shot transfer through multiple instance learning for large images, the text encoder was pre-trained with over 550,000 pathology reports and in-domain text corpora. Evaluated on three WSI datasets from Brigham and Women's Hospital, MI-Zero used independent datasets to prevent information leakage and employed a graph-based representation considering the spatial positions of each patch for slide-level prediction scores, significantly advancing cancer subtype classification accuracy and robustness.

A bi-phase model \cite{arya2021multi} was developed for predicting breast cancer prognosis, integrating multi-modal data from genomic information, histopathology images, and clinical details. The model was evaluated using two datasets: METABRIC (1980 patients) and TCGA-BRCA (1080 patients). The methodology employed a fusion strategy that began with feature extraction from each uni-modal dataset using separate SiGaAtCNNs. The extracted hidden features were then concatenated with input features to form stacked features. These stacked features were subsequently fed into random forest classifiers enhanced with adaptive boosting for the final survival classification task. The fusion strategy leveraged feature-level fusion, combining features from different modalities to augment the model's predictive power, thereby utilizing the strengths of each modality and improving the accuracy of breast cancer prognosis prediction.

\begin{table}[]
\caption{Existing research in multi-modal breast cancer diagnosis}
\label{tab:bc-mm}
\begin{tabular}{lclll}
\hline
\textbf{Author}                                                 & \textbf{Year} & \textbf{Datasets}     & \textbf{\begin{tabular}[c]{@{}l@{}}Fusion\\ strategy\end{tabular}}                                               & \textbf{Modality}                                                                                                          \\ \hline
Sun et al. \cite{sun2018multimodal}           & 2018          & METABRIC              & Late fusion                                                            & \begin{tabular}[c]{@{}l@{}}Clinical data,\\ Gene expression\end{tabular}                                                   \\
Tong et al. \cite{tong2020deep}               & 2020          & TCGA-BRCA             & \begin{tabular}[c]{@{}l@{}}Encoder-decoder \\ method\end{tabular}      & \begin{tabular}[c]{@{}l@{}}Gene Expressions,\\ CNV\end{tabular}                                                            \\
Arya and Saha \cite{arya2021multi}            & 2021          & \begin{tabular}[c]{@{}l@{}}METABRIC,\\ TCGA-BRCA\end{tabular}   & Early fusion                                                           & \begin{tabular}[c]{@{}l@{}}Clinical data, \\ Gene expression\end{tabular}                                                  \\
Subramanian et al. \cite{subramanian2021multi}& 2021          & TCGA-BRCA             & Early fusion                                                           & \begin{tabular}[c]{@{}l@{}}Histopathology images, \\ Clinical data\end{tabular}                                            \\
Liu et al. \cite{liu2022hybrid}               & 2022          & TCGA-BRCA             & Late fusion                                                            & \begin{tabular}[c]{@{}l@{}}Histopathology images, \\ Gene expressions\end{tabular}                                         \\
Howard et al. \cite{howard2022multimodal}     & 2022          & TCGA-BRCA             & Late fusion                                                            & \begin{tabular}[c]{@{}l@{}}Histopathology images, \\ Gene expressions\end{tabular}                                         \\
Arya and Saha \cite{arya2021generative}       & 2022          &\begin{tabular}[c]{@{}l@{}} METABRIC,\\ TCGA-BRCA \end{tabular}   & \begin{tabular}[c]{@{}l@{}}Encoder-decoder \\ method\end{tabular}      & \begin{tabular}[c]{@{}l@{}}Clinical data, \\ Gene expression\end{tabular}                                                  \\
Arya and Saha \cite{arya2020multi}            & 2022          & METABRIC              & Early fusion                                                           & \begin{tabular}[c]{@{}l@{}}Clinical data,\\ Gene expression\end{tabular}                                                   \\
Furtney et al. \cite{furtney2023patient}     & 2023          & TCGA-BRCA             & \begin{tabular}[c]{@{}l@{}}Graph-neural \\ network method\end{tabular} & \begin{tabular}[c]{@{}l@{}}Histopathology images, \\ Clinical data, \\ Gene Expressions, \\ Radiological data\end{tabular} \\
Rani et al. \cite{rani2023diagnosis}          & 2023          & TCGA-BRCA             & Early fusion                                                           & \begin{tabular}[c]{@{}l@{}}Histopathology images, \\ Gene expressions\end{tabular}                                         \\
Kayikci et al. \cite{kayikci2023breast}       & 2023          & METABRIC              & Attention-based                                                        & \begin{tabular}[c]{@{}l@{}}Clinical data, \\ Gene expression\end{tabular}                                                  \\
Arya et al.\cite{arya2023improving}                            & 2023          & TCGA-BRCA             & Early fusion                                                           & \begin{tabular}[c]{@{}l@{}}Clinical data, \\ Gene expression\end{tabular}                                                  \\
Mondol et al. \cite{mondol2024mm}             & 2024          & TCGA-BRCA             & Attention-based                                                        & \begin{tabular}[c]{@{}l@{}}Histopathology images, \\ Clinical data, \\ Gene Expressions\end{tabular}                       \\
Huang et al. \cite{huang2024multimodal}       & 2024          &\begin{tabular}[c]{@{}l@{}} TCGA-BRCA,\\ GMUCH-BRCA \end{tabular}& Early fusion                                                           & \begin{tabular}[c]{@{}l@{}}Histopathology images, \\ Clinical data\end{tabular}                                            \\
Li and Nabavi \cite{li2024multimodal}         & 2024          & TCGA-BRCA             & \begin{tabular}[c]{@{}l@{}}Graph-neural \\ network method\end{tabular} & \begin{tabular}[c]{@{}l@{}}Gene Expressions,\\ CNV\end{tabular}                                                            \\ \hline
\end{tabular}
\end{table}

A hybrid deep learning \cite{liu2022hybrid} model was developed to predict molecular subtypes of breast cancer using gene expression data and pathological images. The TCGA-BRCA dataset, consisting of 1098 samples, was used, with 831 samples selected after filtering. Gene expression data was processed using Log2 and CNVs were detected using Affymetrix SNP 6.0. Pathological images were RGB colored. A multimodal fusion framework was constructed, combining gene and image data with feature extraction networks. The fusion process integrated information from both modalities to enhance the prediction of molecular subtypes. The fusion model outperformed both the DNN and CNN models in terms of accuracy and AUC values, demonstrating the effectiveness of the fusion strategy in improving subtype prediction accuracy.

A deep learning model \cite{howard2022multimodal} was employed to predict recurrence assay results and the risk of recurrence in breast cancer patients. The dataset utilized was The Cancer Genome Atlas (TCGA), comprising 1,099 slides manually annotated to distinguish tumor from surrounding stroma. Tesselated image tiles were extracted from tumor areas and downscaled using a convolutional neural network backbone with fully connected hidden layers for outcome prediction. The model's predictions were based on a combination of digital histology and clinical risk factors, with patient-level predictions calculated by weighting the average of tile-level predictions according to the likelihood of tumor presence in each tile. This fusion strategy led to improved accuracy in predicting recurrence assay results and risk of recurrence, surpassing traditional clinical nomograms. The application of the fusion strategy facilitated enhanced accuracy in predicting recurrence assay results and risk of recurrence in breast cancer patients.

Canonical Correlation Analysis (CCA) and its penalized variants (pCCA) were employed for multi-modality fusion in breast cancer prediction \cite{subramanian2021multi}. CCA identifies correlated linear combinations of two or multiple modalities, while pCCA integrates penalties based on domain knowledge to effectively handle high-dimensional, low-sample-size datasets such as cancer imaging-genomics. The dataset comprised histopathology data and RNA-sequencing data from breast cancer patients in The Cancer Genome Atlas (TCGA). A two-stage prediction pipeline was proposed utilizing pCCA embeddings generated with deflation for latent variable prediction. The fusion strategy amalgamated information from histology and genomics data to augment survival prediction in breast cancer patients. The model demonstrated superior performance compared to Principal Components Analysis (PCA) embeddings in survival prediction tasks.
A deep learning approach was proposed for survival risk stratification in breast cancer, integrating histopathological imaging, genetic, and clinical data \cite{mondol2024mm}. The MaxViT model was employed for image feature extraction, with self-attention mechanisms capturing intricate image relationships at the patient level. A dual cross-attention mechanism fused image features with genetic data to enhance predictive accuracy. The study utilized the TCGA-BRCA dataset, consisting of 249 whole-slide images. Clinical variables such as tumor grade, size, patient age, and lymph node status were meticulously prepared for model evaluations. A dual cross-attention mechanism was utilized to refine the interaction between histopathology image features and genetic expression profiles, employing two attention operations: image features as Query and genetic features as Key and Value, and vice versa.

A Multimodal Deep Neural Network (MDNNMD) was proposed for breast cancer prognosis prediction, incorporating multi-dimensional data \cite{sun2018multimodal}. The method's architecture design and fusion of multi-dimensional data represented innovative aspects. The METABRIC dataset, obtained from 1,980 breast cancer patients, contained gene expression profiles, CNA profiles, and clinical information. Patients were classified as short-term (491) or long-term survivors based on a 5-year survival threshold, with each patient's data comprising 27 clinical features. The MDNNMD method integrated multi-dimensional data, combining gene expression profiles and CNA profiles for comprehensive analysis. The fusion of multi-dimensional data in MDNNMD outperformed single-dimensional data-based methods such as DNN-Clinical, DNN-Expr, and DNN-CNA, highlighting the advantages of this fusion approach.

A two-stage generative incomplete multi-view prediction model, named GIMPP \cite{arya2021generative}, was introduced to address missing view problems in breast cancer prognosis prediction. The first stage utilized multi-view encoder networks and a bi-modal attention scheme to learn common latent space representations. In the second stage, missing view data was generated using view-specific generative adversarial networks conditioned on shared representations and encoded features from other views. The effectiveness of the method was evaluated on the TCGA-BRCA and METABRIC datasets, commonly utilized in cancer research, to demonstrate its superiority over state-of-the-art methods. Fusion in the model was achieved through the generation of missing view data using view-specific generative adversarial networks.
A multimodal siamese model was proposed for breast cancer survival prediction, integrating pathological images and clinical data. Siamese-RegNet was employed to extract survival-related features from pathological patches and capture correlations \cite{huang2024multimodal}. The model comprised patch sampling, feature extraction using Siamese-RegNet, and survival prediction using the LASSO-Cox model. Two datasets, TCGA-BRCA from The Cancer Genome Atlas project and GMUCH-BRCA from Guangzhou Medical University Cancer Hospital, were utilized. The fusion strategy of the model combined information from both images and clinical data to enhance survival prediction accuracy. This comprehensive representation of features facilitated improved survival risk assessment.

A method was utilized to group patient graphs into batches for training and updating graph embeddings in a Graph Neural Network (GNN) framework \cite{furtney2023patient}, employing the cross-entropy loss function. The datasets incorporated in the study comprised The Cancer Genome Atlas Breast Invasive Carcinoma (TCGA-BRCA) dataset, encompassing clinical, genomic, and radiological data from 1,040 patients, with a subset of 108 patients featuring DCE-MRI scans and radiologist tumor measurements. The fusion type adopted in the study involved multimodal fusion of breast cancer patient data with graph convolutional neural networks, modeling patient information into graphs using deep learning features extracted from diverse modalities, such as MRI scans and genomic variant assays. The aim was to enhance the model's generalization potential and performance.


\section{Explainable breast cancer detection: Challenges, trends and future directions}

Explainability remains a critical challenge in breast cancer detection, particularly with the increasing use of complex machine learning and deep learning models. Explainability is essential for clinical decision-making, trust and transparency, regulatory compliance, and error detection and correction. Explainable models enable clinicians to understand the rationale behind a diagnosis, thereby facilitating more informed decision-making and increasing trust in automated systems through transparent conclusions.

\begin{figure}[H]

  \centering
  
  \includegraphics[width=.56\textwidth]{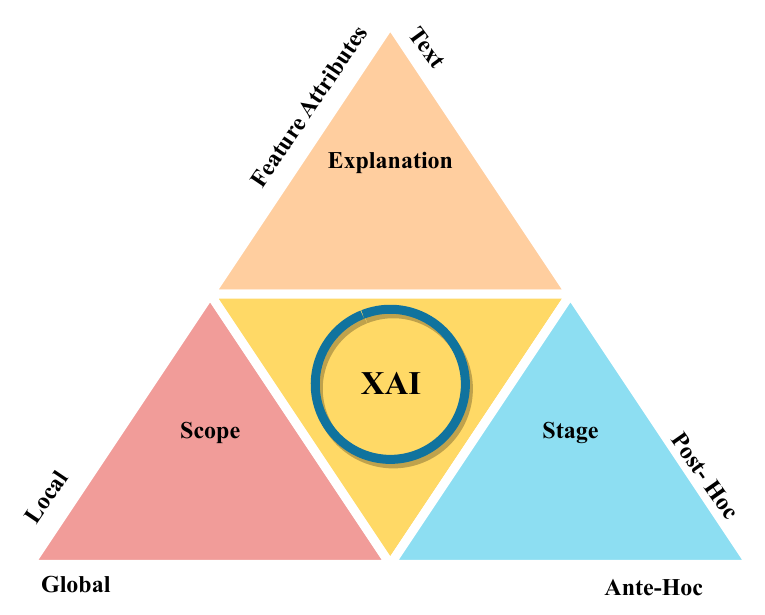}
 
  \caption{Different types of XAI methods }
  \label{fig:xai_meth}
  \end{figure}

However, several challenges in explainability include the complexity of models, data diversity, the black-box nature of algorithms, the trade-off between explainability and accuracy, and the lack of standardization in medical diagnostics. The absence of universally accepted standards for explainability leads to variability in approaches, complicating comparisons and consistent interpretations.

Figure. \ref{fig:xai_meth} illustrates how XAI methods can be categorised in different contexts. Based on Explanation, stage and scope, there can be different methods. Exaplaiability explanations can be in terms of feature attributes and textual format. In scope-based categorisation, there are local and global methods. Post-hoc and ante-hoc are the stage-based XAI methods.

 To enhance explainability, techniques such as feature importance and saliency maps can provide insights into which aspects of the input are driving the model's predictions. Model-agnostic methods like LIME and SHAP allow for the explanation of any machine learning model, offering flexibility in creating explainable outputs. Interpretable models, such as decision trees or linear models, provide greater transparency, albeit potentially at the cost of reduced accuracy. Additionally, human feedback and oversight in the diagnostic process can help validate and explain automated decisions, combining algorithmic efficiency with human intuition. In conclusion, explainability is crucial in breast cancer detection to ensure reliable and trustworthy outcomes, requiring a combination of technical solutions, regulatory compliance, and human oversight to create models that are both accurate and transparent.

\subsection{An explainability analysis on multi-modal techniques}

In the domain of uni-modal breast cancer detection, significant advancements have been made in integrating explainable artificial intelligence (AI) techniques to enhance the interpretability and reliability of predictive models. Gu et al \cite{gu2020case} developed an auxiliary decision support system that combines ensemble learning with case-based reasoning (CBR) to predict breast cancer recurrence. Using XGBoost for predictions and CBR for providing comprehensible explanations, this system effectively communicated the importance of various attributes, aligning well with human reasoning and gaining acceptance among clinicians. Kabakçı et al. \cite{kabakcci2021automated} proposed an automated method for determining CerbB2/HER2 scores from breast tissue images by adhering to ASCO/CAP recommendations. This method employed cell-based image analysis and a hand-crafted feature extraction approach, ensuring both interpretability and adaptability to guideline updates without the need for re-training.

Moreover, recent studies have focused on enhancing the explainability of deep learning models used in breast cancer histopathology. Authors of \cite{maleki2023breast} utilized pre-trained models combined with gradient-boosting classifiers to achieve high accuracy in classifying breast cancer images from the BreakHis dataset. Similarly, Peta et al. \cite{peta2024explainable} introduced an explainable deep learning technique involving adaptive unsharp mask filtering and the Explainable Soft Attentive EfficientNet (ESAE-Net), which provided improved visualization and understanding of classification decisions. Jaume et al. \cite{jaume2020towards} presented CGEXPLAINER, a post-hoc explainer for graph representations in digital pathology, which pruned redundant graph components to maximize mutual information between the original prediction and the sub-graph explanation. These contributions, along with methods like the cost-sensitive CatBoost classifier with LIME explainer \cite{maouche2023explainable} and the use of SHAP for feature importance analysis in tumour cellularity assessment \cite{altini2023tumor}, highlight the growing emphasis on explainability to ensure that AI systems for breast cancer detection are not only accurate but also interpretable and trustworthy for clinical application.

Explainability is a critical factor in radio genomics \cite{liu2022extendable}, as it fosters trust with end-users like physicians and patients, driving the deployment of deep learning models in research and clinical practice. It increases confidence in the model's decision-making process, enabling better understanding and acceptance of results. Explainability also serves as a debugging process for model training and fine-tuning, identifying potential errors or biases. It also helps bypass malicious manipulation, ensuring the integrity and security of radiogenomic research and its applications. In the healthcare field, explainability is especially important as it facilitates better interpretation and understanding of complex AI models, leading to improved patient care and treatment outcomes.

Holzinger et al. \cite{holzinger2021towards} proposed the utilization of Graph Neural Networks (GNNs) as a method for achieving multi-modal causability within explainable AI (xAI). This approach facilitated information fusion through the establishment of causal links between features using graph structures. The method's objective was to construct a multi-modal feature representation space, utilizing knowledge bases as initial connectors for the development of novel explanation interface techniques. Essential components included intra-modal feature extraction and multi-modal embedding. Various GNN architectures and graph embeddings, such as GCNN, Graph Isomorphism Network (GIN), and SchNet, were considered viable options. Additionally, dynamic GNN architectures like Pointer Graph Networks (PGN) were employed to enable the processing of adaptive graphs.
Zhang et al. \cite{zhang2021dmrfnet} introduced a Deep Multimodal Reasoning and Fusion Network (DMRFNet) for Visual Question Answering (VQA) and explanation generation. The model employed multimodal reasoning and fusion techniques to improve the accuracy of answers and explanations. A key innovation was the Multi-Graph Reasoning and Fusion (MGRF) layer, which utilized pre-trained semantic relation embeddings to handle complex spatial and semantic relations among visual objects. DMRFNet was capable of being stacked in depth to facilitate comprehensive reasoning and fusion of multimodal relations. Additionally, an explanation generation module was incorporated to provide justifications for predicted answers. Experimental findings demonstrated the model's effectiveness in achieving both quantitative and qualitative performance improvements.

Authors of \cite{kang2023learning} introduced a segmentation framework with an interpretation module that highlights critical features from each modality, guided by a novel interpretation loss with strengthened and perturbed fusion schemes. This approach effectively generates meaningful interpretable masks, improving multi-modality information integration and segmentation performance. Visualization and perturbation experiments validate the effectiveness of the interpretation method in exploiting meaningful features from each modality.
An interpretable decision-support model for breast cancer diagnosis using histopathology images was proposed in \cite{krishna2023interpretable}. This method integrated an attention branch into a variant of the DarkNet19 CNN model to enhance interpretability and performance. The attention branch generated a heatmap to identify regions of interest, while a perception branch performed image classification through a fully connected layer. Training and validation utilized over 7,000 breast cancer biopsy slide images from the BreaKHis dataset, resulting in a binary classification accuracy of 98.7\%. Notably, the model offered enhanced clinical interpretability, with highlighted cancer regions corresponding well with expert pathologist findings. The ABN-DCN model effectively combined an attention mechanism with a CNN feature extractor, thereby improving both diagnostic interpretability and classification performance in histopathology images.

\section{New frontiers in explainability and  multi-modality}




Recent advancements in multimodality and explainability in medical diagnostics can bring significant improvements in breast cancer detection. These developments highlight potential future directions for integrating advanced computational models and explainability methods into histopathology-based breast cancer diagnostics.
In the context of multimodal methodologies, multimodal fusion is a subset of techniques within the broader field of multimodal analysis. It should be noted that multimodal fusion is prevalent, but that other approaches exist within this field as well. As part of a contemporary focus in multimodal methodologies, image and textual data are integrated, manifesting in applications including report generation \cite{guo2024histgen}, Visual Question Answering (VQA) \cite{hartsock2024vision}, cross-modal retrieval \cite{hu2024histopathology}, and semantic segmentation \cite{van2021hooknet}. It has been noted that substantial scholarly attention has been devoted to the leveraging of medical image and text data through these methodologies\cite{sun2023scoping}. Nevertheless, there still remains a need for further investigation, especially in the context of predicating the diagnosis of breast cancer based on histopathological data using multimodal approaches.

X-VARS \cite{held2024x}, a multimodal large language model initially designed for football refereeing tasks, utilized Video-ChatGPT to process video features and predict responses. This model emphasized interpretability and has demonstrated strong performance in human studies, indicating its potential for adaptation in breast cancer detection. By integrating diverse data sources, such as histopathology images and clinical records, similar models could offer comprehensible diagnostic support, thereby enhancing the accuracy and transparency of the diagnostic process.
The LeGrad \cite{bousselham2024legrad} explainability method, which employs Vision Transformers (ViTs) \cite{vaswani2017attention}, utilizes techniques such as GradCAM \cite{selvaraju2017grad} and AttentionCAM \cite{chen2024actnet} to provide granular insights into feature formation. These explainability methods are crucial for breast cancer detection, offering transparent interpretations of model decisions. By adapting this method to a multimodal scenario that includes histopathology images and clinical/textual data, it can provide comprehensive diagnostic support. This integration enhances trust and clinical applicability by offering transparent and interpretable insights across various data types, thereby improving the accuracy and reliability of breast cancer diagnostics.

The method proposed by Hu et al. \cite{hu2024histopathology}  for fine-grained cross-modal alignment between histopathology WSIs and diagnostic reports holds promise as a future avenue in explainable multimodal breast cancer detection. By leveraging anchor-based WSI and prompt-based text encoders, this method ensured that relevant diagnostic information was accessible and interpretable to pathologists. Through precise alignment and interpretation of multimodal diagnostic data, including histopathology images and clinical and textual reports, the method enhances transparency and interpretability in breast cancer diagnosis. This approach can provide clear insights into the decision-making process of diagnostic models, thereby enhancing trust and clinical acceptance in the application of multimodal AI systems for breast cancer detection.
A multimodal image search strategy was described in \cite{tizhoosh2024image} as a method of improving diagnosis, prognosis, and prediction in histopathology. With this method, large image archives can be explored to identify patterns and correlations using foundation models for feature extraction and image matching. A breast cancer detection framework based on this framework could provide efficient retrieval and comparison of histopathological images, thereby aiding in the identification of malignancies and their characteristics.

Investigating local surrogate explainability techniques in deep learning models, researchers explored the use of VisualBERT and UNITER networks to generate multimodal visual and language explanations \cite{werner2022ability}. The potential of these models to mimic domain expertise underscores the value of explainable AI techniques in breast cancer detection. By providing clear and understandable rationales for automated decisions, such methods enhance clinical trust and support informed decision-making in diagnostic processes.
A framework named LangXAI \cite{nguyen2024langxai} was introduced, integrating explainable AI with advanced vision models to generate textual explanations for visual recognition tasks. This framework enhances transparency and plausibility, potentially improving breast cancer detection by making the diagnostic process more understandable and reliable for clinicians. Consequently, it supports better patient outcomes.

Various explainable AI methods, including Gradient backpropagation and Integrated-Gradients, were applied in \cite{rehman2024envisioning} to analyze the MedCLIP model. These methods provided valuable insights into model predictions, offering pivotal information for the development of breast cancer detection models. Ensuring the transparency and comprehensibility of model decisions can play a crucial role in facilitating regulatory compliance and fostering clinical acceptance of such models in diagnostic settings.
A tool called LVLM-Interpret \cite{ben2024lvlm} was developed to interpret responses from large vision-language models, employing techniques such as raw attention and relevancy maps. This tool's capacity to visualize and comprehend model outputs can be utilized in breast cancer detection to improve the interpretability and reliability of AI-driven diagnostic tools.

An ex-ILP \cite{yang2023neural} framework was introduced to enhance reasoning capabilities in vision-language models by Yang et al. (2023). By improving implicit reasoning skills, this methodology could be harnessed in breast cancer detection to interpret complex interactions between visual and textual data, thus contributing to more accurate and nuanced diagnostic insights. The NLX-GPT method, introduced in \cite{sammani2023uni}, integrated discriminative answer prediction and explanation tasks into a unified model. This approach, which achieves high performance across diverse tasks, holds the potential for adaptation in breast cancer detection. By providing both diagnostic conclusions and their explanations, the NLX-GPT method enhances the usability and trustworthiness of AI models in clinical settings.

These advancements indicate a promising future for the integration of multimodal fusion and explainability in breast cancer diagnostics. By leveraging these innovative approaches, histopathology-based breast cancer diagnosis can be significantly enhanced, leading to improved accuracy, transparency, and ultimately, patient outcomes.

\subsection{Unified Framework: Leveraging Multi-Modal and Explainability}
\begin{figure}[H]

  \centering
  
  \includegraphics[width=.8\textwidth]{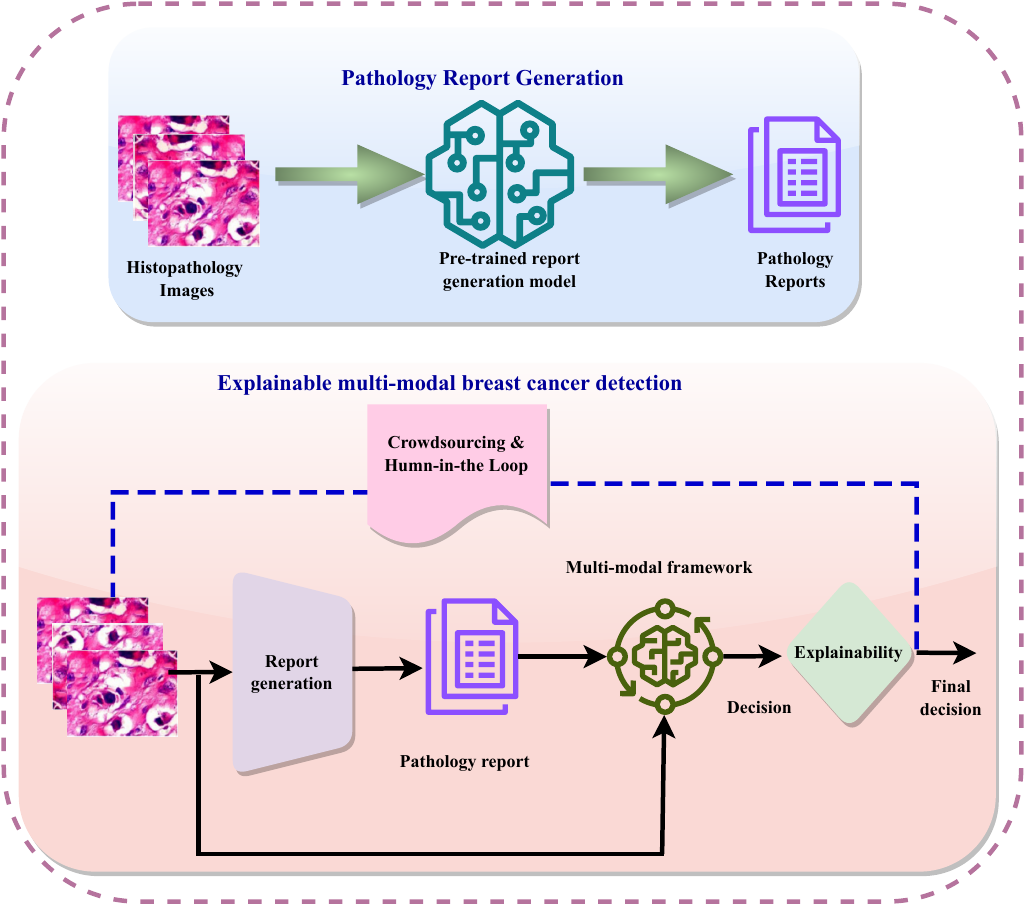}
 
  \caption{Proposed framework for explainable multi-modal breast cancer detection }
  \label{fig:future}
  \end{figure}

The framework proposed for multimodal explainable breast cancer diagnosis involves a systematic process aimed at enhancing diagnostic accuracy and transparency while integrating human expertise for improved patient outcomes. Figure. \ref{fig:future} illustrates the proposed framework.
In the initial step, histopathology images are processed using pre-trained medical report generation models. These models, such as  CLARA \cite{biswal2020clara}, automatically generate comprehensive reports from the images, augmenting them with relevant features extracted through computer vision techniques. Clinical data, including patient history and laboratory results, are integrated into the report generation pipeline to ensure contextually relevant diagnostic reports.

Subsequently, in the multimodal explainable framework for diagnosis, the generated diagnostic reports and histopathology images serve as input. Visual language models, such as Vision Transformers or large language models, are employed to process both visual and textual information simultaneously. Explainability techniques like GradCAM and AttentionCAM are implemented to provide interpretable insights into model decisions, enhancing transparency and trust in the diagnostic process. Model outputs are visualized using tools like LVLM-Interpret to improve interpretability and reliability.

Human expertise is integrated through crowdsourcing or expert consultations to validate and refine model predictions, ensuring clinical relevance and accuracy. This human-in-the-loop approach facilitates informed decision-making and iterative refinement based on feedback from clinicians and patients.
Ultimately, the framework supports diagnostic support by providing transparent and understandable diagnostic conclusions, along with explanations for model predictions. It is integrated into existing clinical workflows to streamline diagnostic processes and enhance patient care, contributing to advancements in the field of breast cancer diagnostics.

\section{Conclusion}
Detection of breast cancer at an early stage and with accuracy is essential to improving patient outcomes. Despite the fact that traditional diagnostic methods primarily rely on single-modal approaches, data analytics combined with diverse sources of data represent a significant shift in diagnostic paradigms. A major step forward in breast cancer diagnosis has been made possible through the adoption of multimodal techniques, which combine both image and non-image data. This review has delved into the burgeoning field of multimodal methodologies, with a specific focus on the fusion of histopathology images with non-image data. Furthermore, the incorporation of Explainable AI (XAI) serves to illuminate the decision-making processes of intricate algorithms, underlining the importance of transparency in diagnostic procedures. By leveraging multi-modal data and emphasizing explainability, this review advocates for enhancing diagnostic accuracy, bolstering clinician confidence, and fostering patient engagement. Moreover, these advancements aim to facilitate the development of personalized treatment strategies tailored to the unique needs of each patient. In addition to addressing the current landscape of multimodality and explainability, this review identifies pertinent research gaps, thereby guiding the trajectory of future studies in this field. By contributing to a comprehensive understanding of multi-modal diagnostic techniques and the imperative of explainability, this review seeks to inform strategic directions in breast cancer diagnosis and treatment, ultimately striving for improved patient outcomes and a more effective healthcare landscape.

\section*{Acknowledgment}
Research reported in this publication was supported by the Qatar Research Development and Innovation Council [ARG01-0513-230141]. The content is solely the responsibility of the authors and does not necessarily represent the official views of Qatar Research Development and Innovation Council.

\bibliography{MM-Hist-XAI-Review}


\begin{thebibliography}{104}
\ifx \bisbn   \undefined \def \bisbn  #1{ISBN #1}\fi
\ifx \binits  \undefined \def \binits#1{#1}\fi
\ifx \bauthor  \undefined \def \bauthor#1{#1}\fi
\ifx \batitle  \undefined \def \batitle#1{#1}\fi
\ifx \bjtitle  \undefined \def \bjtitle#1{#1}\fi
\ifx \bvolume  \undefined \def \bvolume#1{\textbf{#1}}\fi
\ifx \byear  \undefined \def \byear#1{#1}\fi
\ifx \bissue  \undefined \def \bissue#1{#1}\fi
\ifx \bfpage  \undefined \def \bfpage#1{#1}\fi
\ifx \blpage  \undefined \def \blpage #1{#1}\fi
\ifx \burl  \undefined \def \burl#1{\textsf{#1}}\fi
\ifx \doiurl  \undefined \def \doiurl#1{\url{https://doi.org/#1}}\fi
\ifx \betal  \undefined \def \betal{\textit{et al.}}\fi
\ifx \binstitute  \undefined \def \binstitute#1{#1}\fi
\ifx \binstitutionaled  \undefined \def \binstitutionaled#1{#1}\fi
\ifx \bctitle  \undefined \def \bctitle#1{#1}\fi
\ifx \beditor  \undefined \def \beditor#1{#1}\fi
\ifx \bpublisher  \undefined \def \bpublisher#1{#1}\fi
\ifx \bbtitle  \undefined \def \bbtitle#1{#1}\fi
\ifx \bedition  \undefined \def \bedition#1{#1}\fi
\ifx \bseriesno  \undefined \def \bseriesno#1{#1}\fi
\ifx \blocation  \undefined \def \blocation#1{#1}\fi
\ifx \bsertitle  \undefined \def \bsertitle#1{#1}\fi
\ifx \bsnm \undefined \def \bsnm#1{#1}\fi
\ifx \bsuffix \undefined \def \bsuffix#1{#1}\fi
\ifx \bparticle \undefined \def \bparticle#1{#1}\fi
\ifx \barticle \undefined \def \barticle#1{#1}\fi
\bibcommenthead
\ifx \bconfdate \undefined \def \bconfdate #1{#1}\fi
\ifx \botherref \undefined \def \botherref #1{#1}\fi
\ifx \url \undefined \def \url#1{\textsf{#1}}\fi
\ifx \bchapter \undefined \def \bchapter#1{#1}\fi
\ifx \bbook \undefined \def \bbook#1{#1}\fi
\ifx \bcomment \undefined \def \bcomment#1{#1}\fi
\ifx \oauthor \undefined \def \oauthor#1{#1}\fi
\ifx \citeauthoryear \undefined \def \citeauthoryear#1{#1}\fi
\ifx \endbibitem  \undefined \def \endbibitem {}\fi
\ifx \bconflocation  \undefined \def \bconflocation#1{#1}\fi
\ifx \arxivurl  \undefined \def \arxivurl#1{\textsf{#1}}\fi
\csname PreBibitemsHook\endcsname

\bibitem[\protect\citeauthoryear{Sun et~al.}{2023}]{sun2023scoping}
\begin{botherref}
\oauthor{\bsnm{Sun}, \binits{Z.}},
\oauthor{\bsnm{Lin}, \binits{M.}},
\oauthor{\bsnm{Zhu}, \binits{Q.}},
\oauthor{\bsnm{Xie}, \binits{Q.}},
\oauthor{\bsnm{Wang}, \binits{F.}},
\oauthor{\bsnm{Lu}, \binits{Z.}},
\oauthor{\bsnm{Peng}, \binits{Y.}}:
A scoping review on multimodal deep learning in biomedical images and texts.
Journal of Biomedical Informatics,
104482
(2023)
\end{botherref}
\endbibitem

\bibitem[\protect\citeauthoryear{Krithiga and Geetha}{2021}]{krithiga2021breast}
\begin{barticle}
\bauthor{\bsnm{Krithiga}, \binits{R.}},
\bauthor{\bsnm{Geetha}, \binits{P.}}:
\batitle{Breast cancer detection, segmentation and classification on histopathology images analysis: a systematic review}.
\bjtitle{Archives of Computational Methods in Engineering}
\bvolume{28}(\bissue{4}),
\bfpage{2607}--\blpage{2619}
(\byear{2021})
\end{barticle}
\endbibitem

\bibitem[\protect\citeauthoryear{Tafavvoghi et~al.}{2024}]{tafavvoghi2024publicly}
\begin{botherref}
\oauthor{\bsnm{Tafavvoghi}, \binits{M.}},
\oauthor{\bsnm{Bongo}, \binits{L.A.}},
\oauthor{\bsnm{Shvetsov}, \binits{N.}},
\oauthor{\bsnm{Busund}, \binits{L.-T.R.}},
\oauthor{\bsnm{M{\o}llersen}, \binits{K.}}:
Publicly available datasets of breast histopathology h\&e whole-slide images: A scoping review.
Journal of Pathology Informatics,
100363
(2024)
\end{botherref}
\endbibitem

\bibitem[\protect\citeauthoryear{Abo-El-Rejal et~al.}{2024}]{abo2024advances}
\begin{barticle}
\bauthor{\bsnm{Abo-El-Rejal}, \binits{A.}},
\bauthor{\bsnm{Ayman}, \binits{S.}},
\bauthor{\bsnm{Aymen}, \binits{F.}}:
\batitle{Advances in breast cancer segmentation: A comprehensive review}.
\bjtitle{Acadlore Transactions on AI and Machine Learning}
\bvolume{3}(\bissue{2}),
\bfpage{70}--\blpage{83}
(\byear{2024})
\end{barticle}
\endbibitem

\bibitem[\protect\citeauthoryear{Hussain et~al.}{2024}]{hussain2024breast}
\begin{barticle}
\bauthor{\bsnm{Hussain}, \binits{S.}},
\bauthor{\bsnm{Ali}, \binits{M.}},
\bauthor{\bsnm{Naseem}, \binits{U.}},
\bauthor{\bsnm{Nezhadmoghadam}, \binits{F.}},
\bauthor{\bsnm{Jatoi}, \binits{M.A.}},
\bauthor{\bsnm{Gulliver}, \binits{T.A.}},
\bauthor{\bsnm{Tamez-Pe{\~n}a}, \binits{J.G.}}:
\batitle{Breast cancer risk prediction using machine learning: a systematic review}.
\bjtitle{Frontiers in Oncology}
\bvolume{14},
\bfpage{1343627}
(\byear{2024})
\end{barticle}
\endbibitem

\bibitem[\protect\citeauthoryear{Brodhead et~al.}{2024}]{brodhead2024multimodality}
\begin{botherref}
\oauthor{\bsnm{Brodhead}, \binits{M.}},
\oauthor{\bsnm{Woods}, \binits{R.W.}},
\oauthor{\bsnm{Fowler}, \binits{A.M.}},
\oauthor{\bsnm{Roy}, \binits{M.}},
\oauthor{\bsnm{Neuman}, \binits{H.}},
\oauthor{\bsnm{Gegios}, \binits{A.}}:
Multimodality imaging review of metastatic melanoma involving the breast.
Current Problems in Diagnostic Radiology
(2024)
\end{botherref}
\endbibitem

\bibitem[\protect\citeauthoryear{Luo et~al.}{2024}]{luo2024deep}
\begin{botherref}
\oauthor{\bsnm{Luo}, \binits{L.}},
\oauthor{\bsnm{Wang}, \binits{X.}},
\oauthor{\bsnm{Lin}, \binits{Y.}},
\oauthor{\bsnm{Ma}, \binits{X.}},
\oauthor{\bsnm{Tan}, \binits{A.}},
\oauthor{\bsnm{Chan}, \binits{R.}},
\oauthor{\bsnm{Vardhanabhuti}, \binits{V.}},
\oauthor{\bsnm{Chu}, \binits{W.C.}},
\oauthor{\bsnm{Cheng}, \binits{K.-T.}},
\oauthor{\bsnm{Chen}, \binits{H.}}:
Deep learning in breast cancer imaging: A decade of progress and future directions.
IEEE Reviews in Biomedical Engineering
(2024)
\end{botherref}
\endbibitem

\bibitem[\protect\citeauthoryear{Rautela et~al.}{2024}]{rautela2024comprehensive}
\begin{botherref}
\oauthor{\bsnm{Rautela}, \binits{K.}},
\oauthor{\bsnm{Kumar}, \binits{D.}},
\oauthor{\bsnm{Kumar}, \binits{V.}}:
A comprehensive review on computational techniques for breast cancer: past, present, and future.
Multimedia Tools and Applications,
1--34
(2024)
\end{botherref}
\endbibitem

\bibitem[\protect\citeauthoryear{Singh et~al.}{2024}]{singh2024technical}
\begin{bchapter}
\bauthor{\bsnm{Singh}, \binits{A.}},
\bauthor{\bsnm{Kaur}, \binits{S.}},
\bauthor{\bsnm{Singh}, \binits{D.}},
\bauthor{\bsnm{Singh}, \binits{G.}}:
\bctitle{Technical review of breast cancer screening and detection using artificial intelligence and radiomics}.
In: \bbtitle{2024 11th International Conference on Computing for Sustainable Global Development (INDIACom)},
pp. \bfpage{1171}--\blpage{1176}
(\byear{2024}).
\bcomment{IEEE}
\end{bchapter}
\endbibitem

\bibitem[\protect\citeauthoryear{Thakur et~al.}{2024}]{thakur2024systematic}
\begin{barticle}
\bauthor{\bsnm{Thakur}, \binits{N.}},
\bauthor{\bsnm{Kumar}, \binits{P.}},
\bauthor{\bsnm{Kumar}, \binits{A.}}:
\batitle{A systematic review of machine and deep learning techniques for the identification and classification of breast cancer through medical image modalities}.
\bjtitle{Multimedia Tools and Applications}
\bvolume{83}(\bissue{12}),
\bfpage{35849}--\blpage{35942}
(\byear{2024})
\end{barticle}
\endbibitem

\bibitem[\protect\citeauthoryear{Nasser and Yusof}{2023}]{nasser2023deep}
\begin{barticle}
\bauthor{\bsnm{Nasser}, \binits{M.}},
\bauthor{\bsnm{Yusof}, \binits{U.K.}}:
\batitle{Deep learning based methods for breast cancer diagnosis: a systematic review and future direction}.
\bjtitle{Diagnostics}
\bvolume{13}(\bissue{1}),
\bfpage{161}
(\byear{2023})
\end{barticle}
\endbibitem

\bibitem[\protect\citeauthoryear{Obeagu and Obeagu}{2024}]{obeagu2024breast}
\begin{barticle}
\bauthor{\bsnm{Obeagu}, \binits{E.I.}},
\bauthor{\bsnm{Obeagu}, \binits{G.U.}}:
\batitle{Breast cancer: A review of risk factors and diagnosis}.
\bjtitle{Medicine}
\bvolume{103}(\bissue{3}),
\bfpage{36905}
(\byear{2024})
\end{barticle}
\endbibitem

\bibitem[\protect\citeauthoryear{Rai}{2024}]{rai2024cancer}
\begin{barticle}
\bauthor{\bsnm{Rai}, \binits{H.M.}}:
\batitle{Cancer detection and segmentation using machine learning and deep learning techniques: A review}.
\bjtitle{Multimedia Tools and Applications}
\bvolume{83}(\bissue{9}),
\bfpage{27001}--\blpage{27035}
(\byear{2024})
\end{barticle}
\endbibitem

\bibitem[\protect\citeauthoryear{Liu et~al.}{2023}]{liu2023classifier}
\begin{barticle}
\bauthor{\bsnm{Liu}, \binits{Z.}},
\bauthor{\bsnm{Lin}, \binits{F.}},
\bauthor{\bsnm{Huang}, \binits{J.}},
\bauthor{\bsnm{Wu}, \binits{X.}},
\bauthor{\bsnm{Wen}, \binits{J.}},
\bauthor{\bsnm{Wang}, \binits{M.}},
\bauthor{\bsnm{Ren}, \binits{Y.}},
\bauthor{\bsnm{Wei}, \binits{X.}},
\bauthor{\bsnm{Song}, \binits{X.}},
\bauthor{\bsnm{Qin}, \binits{J.}}, \betal:
\batitle{A classifier-combined method for grading breast cancer based on dempster-shafer evidence theory}.
\bjtitle{Quantitative Imaging in Medicine and Surgery}
\bvolume{13}(\bissue{5}),
\bfpage{3288}
(\byear{2023})
\end{barticle}
\endbibitem

\bibitem[\protect\citeauthoryear{Kumaraswamy et~al.}{2023}]{kumaraswamy2023invasive}
\begin{barticle}
\bauthor{\bsnm{Kumaraswamy}, \binits{E.}},
\bauthor{\bsnm{Kumar}, \binits{S.}},
\bauthor{\bsnm{Sharma}, \binits{M.}}:
\batitle{An invasive ductal carcinomas breast cancer grade classification using an ensemble of convolutional neural networks}.
\bjtitle{Diagnostics}
\bvolume{13}(\bissue{11}),
\bfpage{1977}
(\byear{2023})
\end{barticle}
\endbibitem

\bibitem[\protect\citeauthoryear{Huang et~al.}{2024}]{huang2024classifying}
\begin{barticle}
\bauthor{\bsnm{Huang}, \binits{Y.}},
\bauthor{\bsnm{Zeng}, \binits{P.}},
\bauthor{\bsnm{Zhong}, \binits{C.}}:
\batitle{Classifying breast cancer subtypes on multi-omics data via sparse canonical correlation analysis and deep learning}.
\bjtitle{BMC bioinformatics}
\bvolume{25}(\bissue{1}),
\bfpage{132}
(\byear{2024})
\end{barticle}
\endbibitem

\bibitem[\protect\citeauthoryear{Choi and Chae}{2023}]{choi2023mobrca}
\begin{barticle}
\bauthor{\bsnm{Choi}, \binits{J.M.}},
\bauthor{\bsnm{Chae}, \binits{H.}}:
\batitle{mobrca-net: a breast cancer subtype classification framework based on multi-omics attention neural networks}.
\bjtitle{BMC bioinformatics}
\bvolume{24}(\bissue{1}),
\bfpage{169}
(\byear{2023})
\end{barticle}
\endbibitem

\bibitem[\protect\citeauthoryear{Raza et~al.}{2023}]{raza2023deepbreastcancernet}
\begin{barticle}
\bauthor{\bsnm{Raza}, \binits{A.}},
\bauthor{\bsnm{Ullah}, \binits{N.}},
\bauthor{\bsnm{Khan}, \binits{J.A.}},
\bauthor{\bsnm{Assam}, \binits{M.}},
\bauthor{\bsnm{Guzzo}, \binits{A.}},
\bauthor{\bsnm{Aljuaid}, \binits{H.}}:
\batitle{Deepbreastcancernet: A novel deep learning model for breast cancer detection using ultrasound images}.
\bjtitle{Applied Sciences}
\bvolume{13}(\bissue{4}),
\bfpage{2082}
(\byear{2023})
\end{barticle}
\endbibitem

\bibitem[\protect\citeauthoryear{Al-Dhabyani et~al.}{2020}]{al2020dataset}
\begin{barticle}
\bauthor{\bsnm{Al-Dhabyani}, \binits{W.}},
\bauthor{\bsnm{Gomaa}, \binits{M.}},
\bauthor{\bsnm{Khaled}, \binits{H.}},
\bauthor{\bsnm{Fahmy}, \binits{A.}}:
\batitle{Dataset of breast ultrasound images}.
\bjtitle{Data in brief}
\bvolume{28},
\bfpage{104863}
(\byear{2020})
\end{barticle}
\endbibitem

\bibitem[\protect\citeauthoryear{Paulo}{2017}]{paulo2017breast}
\begin{botherref}
\oauthor{\bsnm{Paulo}, \binits{S.}}:
Breast ultrasound image. Mendeley data
(2017)
\end{botherref}
\endbibitem

\bibitem[\protect\citeauthoryear{{The Cancer Genome Atlas (TCGA)}}{}]{tcga2023}
\begin{botherref}
\oauthor{\bsnm{{The Cancer Genome Atlas (TCGA)}}}:
Genomic Data Commons Data Portal (GDC).
\url{https://portal.gdc.cancer.gov/projects/TCGA-BRCA}.
Accessed 07 Jul. 2023
\end{botherref}
\endbibitem

\bibitem[\protect\citeauthoryear{Parshionikar and Bhattacharyya}{2024}]{parshionikar2024enhanced}
\begin{barticle}
\bauthor{\bsnm{Parshionikar}, \binits{S.}},
\bauthor{\bsnm{Bhattacharyya}, \binits{D.}}:
\batitle{An enhanced multi-scale deep convolutional orchard capsule neural network for multi-modal breast cancer detection}.
\bjtitle{Healthcare Analytics}
\bvolume{5},
\bfpage{100298}
(\byear{2024})
\end{barticle}
\endbibitem

\bibitem[\protect\citeauthoryear{Spanhol et~al.}{2015}]{spanhol2015dataset}
\begin{barticle}
\bauthor{\bsnm{Spanhol}, \binits{F.A.}},
\bauthor{\bsnm{Oliveira}, \binits{L.S.}},
\bauthor{\bsnm{Petitjean}, \binits{C.}},
\bauthor{\bsnm{Heutte}, \binits{L.}}:
\batitle{A dataset for breast cancer histopathological image classification}.
\bjtitle{Ieee transactions on biomedical engineering}
\bvolume{63}(\bissue{7}),
\bfpage{1455}--\blpage{1462}
(\byear{2015})
\end{barticle}
\endbibitem

\bibitem[\protect\citeauthoryear{Zuluaga-Gomez et~al.}{2021}]{zuluaga2021cnn}
\begin{barticle}
\bauthor{\bsnm{Zuluaga-Gomez}, \binits{J.}},
\bauthor{\bsnm{Al~Masry}, \binits{Z.}},
\bauthor{\bsnm{Benaggoune}, \binits{K.}},
\bauthor{\bsnm{Meraghni}, \binits{S.}},
\bauthor{\bsnm{Zerhouni}, \binits{N.}}:
\batitle{A cnn-based methodology for breast cancer diagnosis using thermal images}.
\bjtitle{Computer Methods in Biomechanics and Biomedical Engineering: Imaging \& Visualization}
\bvolume{9}(\bissue{2}),
\bfpage{131}--\blpage{145}
(\byear{2021})
\end{barticle}
\endbibitem

\bibitem[\protect\citeauthoryear{{DataBioX}}{2024}]{databiox_datasets}
\begin{botherref}
\oauthor{\bsnm{{DataBioX}}}:
DataBioX Datasets.
Accessed: 2024-06-02
(2024).
\url{https://databiox.com/datasets/}
\end{botherref}
\endbibitem

\bibitem[\protect\citeauthoryear{Sahu et~al.}{2023a}]{sahu2023cnn}
\begin{barticle}
\bauthor{\bsnm{Sahu}, \binits{Y.}},
\bauthor{\bsnm{Tripathi}, \binits{A.}},
\bauthor{\bsnm{Gupta}, \binits{R.K.}},
\bauthor{\bsnm{Gautam}, \binits{P.}},
\bauthor{\bsnm{Pateriya}, \binits{R.K.}},
\bauthor{\bsnm{Gupta}, \binits{A.}}:
\batitle{A cnn-svm based computer aided diagnosis of breast cancer using histogram k-means segmentation technique}.
\bjtitle{Multimedia Tools and Applications}
\bvolume{82}(\bissue{9}),
\bfpage{14055}--\blpage{14075}
(\byear{2023})
\end{barticle}
\endbibitem

\bibitem[\protect\citeauthoryear{Sahu et~al.}{2023b}]{sahu2023high}
\begin{barticle}
\bauthor{\bsnm{Sahu}, \binits{A.}},
\bauthor{\bsnm{Das}, \binits{P.K.}},
\bauthor{\bsnm{Meher}, \binits{S.}}:
\batitle{High accuracy hybrid cnn classifiers for breast cancer detection using mammogram and ultrasound datasets}.
\bjtitle{Biomedical Signal Processing and Control}
\bvolume{80},
\bfpage{104292}
(\byear{2023})
\end{barticle}
\endbibitem

\bibitem[\protect\citeauthoryear{Lekamlage et~al.}{2020}]{lekamlage2020mini}
\begin{bchapter}
\bauthor{\bsnm{Lekamlage}, \binits{C.D.}},
\bauthor{\bsnm{Afzal}, \binits{F.}},
\bauthor{\bsnm{Westerberg}, \binits{E.}},
\bauthor{\bsnm{Cheddad}, \binits{A.}}:
\bctitle{Mini-ddsm: Mammography-based automatic age estimation}.
In: \bbtitle{2020 3rd International Conference on Digital Medicine and Image Processing},
pp. \bfpage{1}--\blpage{6}
(\byear{2020})
\end{bchapter}
\endbibitem

\bibitem[\protect\citeauthoryear{Srikantamurthy et~al.}{2023}]{srikantamurthy2023classification}
\begin{barticle}
\bauthor{\bsnm{Srikantamurthy}, \binits{M.M.}},
\bauthor{\bsnm{Rallabandi}, \binits{V.S.}},
\bauthor{\bsnm{Dudekula}, \binits{D.B.}},
\bauthor{\bsnm{Natarajan}, \binits{S.}},
\bauthor{\bsnm{Park}, \binits{J.}}:
\batitle{Classification of benign and malignant subtypes of breast cancer histopathology imaging using hybrid cnn-lstm based transfer learning}.
\bjtitle{BMC Medical Imaging}
\bvolume{23}(\bissue{1}),
\bfpage{19}
(\byear{2023})
\end{barticle}
\endbibitem

\bibitem[\protect\citeauthoryear{Guo et~al.}{2024}]{guo2024multimodal}
\begin{barticle}
\bauthor{\bsnm{Guo}, \binits{D.}},
\bauthor{\bsnm{Lu}, \binits{C.}},
\bauthor{\bsnm{Chen}, \binits{D.}},
\bauthor{\bsnm{Yuan}, \binits{J.}},
\bauthor{\bsnm{Duan}, \binits{Q.}},
\bauthor{\bsnm{Xue}, \binits{Z.}},
\bauthor{\bsnm{Liu}, \binits{S.}},
\bauthor{\bsnm{Huang}, \binits{Y.}}:
\batitle{A multimodal breast cancer diagnosis method based on knowledge-augmented deep learning}.
\bjtitle{Biomedical Signal Processing and Control}
\bvolume{90},
\bfpage{105843}
(\byear{2024})
\end{barticle}
\endbibitem

\bibitem[\protect\citeauthoryear{Liu et~al.}{2024}]{liu2024multi}
\begin{barticle}
\bauthor{\bsnm{Liu}, \binits{H.}},
\bauthor{\bsnm{Shi}, \binits{Y.}},
\bauthor{\bsnm{Li}, \binits{A.}},
\bauthor{\bsnm{Wang}, \binits{M.}}:
\batitle{Multi-modal fusion network with intra-and inter-modality attention for prognosis prediction in breast cancer}.
\bjtitle{Computers in Biology and Medicine}
\bvolume{168},
\bfpage{107796}
(\byear{2024})
\end{barticle}
\endbibitem

\bibitem[\protect\citeauthoryear{Sivamurugan and Sureshkumar}{2023}]{sivamurugan2023applying}
\begin{barticle}
\bauthor{\bsnm{Sivamurugan}, \binits{J.}},
\bauthor{\bsnm{Sureshkumar}, \binits{G.}}:
\batitle{Applying dual models on optimized lstm with u-net segmentation for breast cancer diagnosis using mammogram images}.
\bjtitle{Artificial Intelligence in Medicine}
\bvolume{143},
\bfpage{102626}
(\byear{2023})
\end{barticle}
\endbibitem

\bibitem[\protect\citeauthoryear{Kendall et~al.}{2013}]{kendall2013automatic}
\begin{barticle}
\bauthor{\bsnm{Kendall}, \binits{E.J.}},
\bauthor{\bsnm{Barnett}, \binits{M.G.}},
\bauthor{\bsnm{Chytyk-Praznik}, \binits{K.}}:
\batitle{Automatic detection of anomalies in screening mammograms}.
\bjtitle{BMC Medical Imaging}
\bvolume{13},
\bfpage{1}--\blpage{11}
(\byear{2013})
\end{barticle}
\endbibitem

\bibitem[\protect\citeauthoryear{Murata et~al.}{2023}]{murata2023prediction}
\begin{barticle}
\bauthor{\bsnm{Murata}, \binits{T.}},
\bauthor{\bsnm{Yoshida}, \binits{M.}},
\bauthor{\bsnm{Shiino}, \binits{S.}},
\bauthor{\bsnm{Ogawa}, \binits{A.}},
\bauthor{\bsnm{Watase}, \binits{C.}},
\bauthor{\bsnm{Satomi}, \binits{K.}},
\bauthor{\bsnm{Jimbo}, \binits{K.}},
\bauthor{\bsnm{Maeshima}, \binits{A.}},
\bauthor{\bsnm{Iwamoto}, \binits{E.}},
\bauthor{\bsnm{Takayama}, \binits{S.}}, \betal:
\batitle{A prediction model for distant metastasis after isolated locoregional recurrence of breast cancer}.
\bjtitle{Breast Cancer Research and Treatment}
\bvolume{199}(\bissue{1}),
\bfpage{57}--\blpage{66}
(\byear{2023})
\end{barticle}
\endbibitem

\bibitem[\protect\citeauthoryear{Hussein et~al.}{2024}]{hussein2024framework}
\begin{barticle}
\bauthor{\bsnm{Hussein}, \binits{M.}},
\bauthor{\bsnm{Elnahas}, \binits{M.}},
\bauthor{\bsnm{Keshk}, \binits{A.}}:
\batitle{A framework for predicting breast cancer recurrence}.
\bjtitle{Expert Systems with Applications}
\bvolume{240},
\bfpage{122641}
(\byear{2024})
\end{barticle}
\endbibitem

\bibitem[\protect\citeauthoryear{Ahmed et~al.}{2023}]{ahmed2023images}
\begin{barticle}
\bauthor{\bsnm{Ahmed}, \binits{L.}},
\bauthor{\bsnm{Iqbal}, \binits{M.M.}},
\bauthor{\bsnm{Aldabbas}, \binits{H.}},
\bauthor{\bsnm{Khalid}, \binits{S.}},
\bauthor{\bsnm{Saleem}, \binits{Y.}},
\bauthor{\bsnm{Saeed}, \binits{S.}}:
\batitle{Images data practices for semantic segmentation of breast cancer using deep neural network}.
\bjtitle{Journal of Ambient Intelligence and Humanized Computing}
\bvolume{14}(\bissue{11}),
\bfpage{15227}--\blpage{15243}
(\byear{2023})
\end{barticle}
\endbibitem

\bibitem[\protect\citeauthoryear{Lee et~al.}{2017}]{lee2017curated}
\begin{barticle}
\bauthor{\bsnm{Lee}, \binits{R.S.}},
\bauthor{\bsnm{Gimenez}, \binits{F.}},
\bauthor{\bsnm{Hoogi}, \binits{A.}},
\bauthor{\bsnm{Miyake}, \binits{K.K.}},
\bauthor{\bsnm{Gorovoy}, \binits{M.}},
\bauthor{\bsnm{Rubin}, \binits{D.L.}}:
\batitle{A curated mammography data set for use in computer-aided detection and diagnosis research}.
\bjtitle{Scientific data}
\bvolume{4}(\bissue{1}),
\bfpage{1}--\blpage{9}
(\byear{2017})
\end{barticle}
\endbibitem

\bibitem[\protect\citeauthoryear{Alam et~al.}{2023}]{alam2023improving}
\begin{barticle}
\bauthor{\bsnm{Alam}, \binits{T.}},
\bauthor{\bsnm{Shia}, \binits{W.-C.}},
\bauthor{\bsnm{Hsu}, \binits{F.-R.}},
\bauthor{\bsnm{Hassan}, \binits{T.}}:
\batitle{Improving breast cancer detection and diagnosis through semantic segmentation using the unet3+ deep learning framework}.
\bjtitle{Biomedicines}
\bvolume{11}(\bissue{6}),
\bfpage{1536}
(\byear{2023})
\end{barticle}
\endbibitem

\bibitem[\protect\citeauthoryear{Prinzi et~al.}{2024}]{prinzi2024yolo}
\begin{barticle}
\bauthor{\bsnm{Prinzi}, \binits{F.}},
\bauthor{\bsnm{Insalaco}, \binits{M.}},
\bauthor{\bsnm{Orlando}, \binits{A.}},
\bauthor{\bsnm{Gaglio}, \binits{S.}},
\bauthor{\bsnm{Vitabile}, \binits{S.}}:
\batitle{A yolo-based model for breast cancer detection in mammograms}.
\bjtitle{Cognitive Computation}
\bvolume{16}(\bissue{1}),
\bfpage{107}--\blpage{120}
(\byear{2024})
\end{barticle}
\endbibitem

\bibitem[\protect\citeauthoryear{Moreira et~al.}{2012}]{moreira2012inbreast}
\begin{barticle}
\bauthor{\bsnm{Moreira}, \binits{I.C.}},
\bauthor{\bsnm{Amaral}, \binits{I.}},
\bauthor{\bsnm{Domingues}, \binits{I.}},
\bauthor{\bsnm{Cardoso}, \binits{A.}},
\bauthor{\bsnm{Cardoso}, \binits{M.J.}},
\bauthor{\bsnm{Cardoso}, \binits{J.S.}}:
\batitle{Inbreast: toward a full-field digital mammographic database}.
\bjtitle{Academic radiology}
\bvolume{19}(\bissue{2}),
\bfpage{236}--\blpage{248}
(\byear{2012})
\end{barticle}
\endbibitem

\bibitem[\protect\citeauthoryear{Guo et~al.}{2024}]{guo2024multi}
\begin{barticle}
\bauthor{\bsnm{Guo}, \binits{H.}},
\bauthor{\bsnm{Li}, \binits{M.}},
\bauthor{\bsnm{Liu}, \binits{H.}},
\bauthor{\bsnm{Chen}, \binits{X.}},
\bauthor{\bsnm{Cheng}, \binits{Z.}},
\bauthor{\bsnm{Li}, \binits{X.}},
\bauthor{\bsnm{Yu}, \binits{H.}},
\bauthor{\bsnm{He}, \binits{Q.}}:
\batitle{Multi-threshold image segmentation based on an improved salp swarm algorithm: Case study of breast cancer pathology images}.
\bjtitle{Computers in Biology and Medicine}
\bvolume{168},
\bfpage{107769}
(\byear{2024})
\end{barticle}
\endbibitem

\bibitem[\protect\citeauthoryear{Rajoub et~al.}{2024}]{rajoub2024segmentation}
\begin{botherref}
\oauthor{\bsnm{Rajoub}, \binits{B.}},
\oauthor{\bsnm{Qusa}, \binits{H.}},
\oauthor{\bsnm{Abdul-Rahman}, \binits{H.}},
\oauthor{\bsnm{Mohamed}, \binits{H.}}:
Segmentation of breast tissue structures in mammographic images.
Artificial Intelligence and Image Processing in Medical Imaging,
115--146
(2024)
\end{botherref}
\endbibitem

\bibitem[\protect\citeauthoryear{Soliman et~al.}{2024}]{soliman2024artificial}
\begin{barticle}
\bauthor{\bsnm{Soliman}, \binits{A.}},
\bauthor{\bsnm{Li}, \binits{Z.}},
\bauthor{\bsnm{Parwani}, \binits{A.V.}}:
\batitle{Artificial intelligence’s impact on breast cancer pathology: a literature review}.
\bjtitle{Diagnostic Pathology}
\bvolume{19}(\bissue{1}),
\bfpage{1}--\blpage{18}
(\byear{2024})
\end{barticle}
\endbibitem

\bibitem[\protect\citeauthoryear{Gallagher et~al.}{2024}]{gallagher2024artificial}
\begin{botherref}
\oauthor{\bsnm{Gallagher}, \binits{W.M.}},
\oauthor{\bsnm{McCaffrey}, \binits{C.}},
\oauthor{\bsnm{Jahangir}, \binits{C.}},
\oauthor{\bsnm{Murphy}, \binits{C.}},
\oauthor{\bsnm{Burke}, \binits{C.}},
\oauthor{\bsnm{Rahman}, \binits{A.}}:
Artificial intelligence in digital histopathology for predicting patient prognosis and treatment efficacy in breast cancer.
Expert Review of Molecular Diagnostics
(just-accepted)
(2024)
\end{botherref}
\endbibitem

\bibitem[\protect\citeauthoryear{Sweetlin and Saudia}{2021}]{sweetlin2021review}
\begin{bchapter}
\bauthor{\bsnm{Sweetlin}, \binits{E.J.}},
\bauthor{\bsnm{Saudia}, \binits{S.}}:
\bctitle{A review of machine learning algorithms on different breast cancer datasets}.
In: \bbtitle{International Conference on Big Data, Machine Learning, and Applications},
pp. \bfpage{659}--\blpage{673}
(\byear{2021}).
\bcomment{Springer}
\end{bchapter}
\endbibitem

\bibitem[\protect\citeauthoryear{Heiliger et~al.}{2023}]{heiliger2023beyond}
\begin{botherref}
\oauthor{\bsnm{Heiliger}, \binits{L.}},
\oauthor{\bsnm{Sekuboyina}, \binits{A.}},
\oauthor{\bsnm{Menze}, \binits{B.}},
\oauthor{\bsnm{Egger}, \binits{J.}},
\oauthor{\bsnm{Kleesiek}, \binits{J.}}:
Beyond medical imaging-a review of multimodal deep learning in radiology.
Authorea Preprints
(2023)
\end{botherref}
\endbibitem

\bibitem[\protect\citeauthoryear{Laokulrath et~al.}{2024}]{laokulrath2024invasive}
\begin{botherref}
\oauthor{\bsnm{Laokulrath}, \binits{N.}},
\oauthor{\bsnm{Gudi}, \binits{M.A.}},
\oauthor{\bsnm{Deb}, \binits{R.}},
\oauthor{\bsnm{Ellis}, \binits{I.O.}},
\oauthor{\bsnm{Tan}, \binits{P.H.}}:
Invasive breast cancer reporting guidelines: Iccr, cap, rcpath, rcpa datasets and future directions.
Diagnostic Histopathology
(2024)
\end{botherref}
\endbibitem

\bibitem[\protect\citeauthoryear{Brancati et~al.}{2022}]{brancati2022bracs}
\begin{barticle}
\bauthor{\bsnm{Brancati}, \binits{N.}},
\bauthor{\bsnm{Anniciello}, \binits{A.M.}},
\bauthor{\bsnm{Pati}, \binits{P.}},
\bauthor{\bsnm{Riccio}, \binits{D.}},
\bauthor{\bsnm{Scognamiglio}, \binits{G.}},
\bauthor{\bsnm{Jaume}, \binits{G.}},
\bauthor{\bsnm{De~Pietro}, \binits{G.}},
\bauthor{\bsnm{Di~Bonito}, \binits{M.}},
\bauthor{\bsnm{Foncubierta}, \binits{A.}},
\bauthor{\bsnm{Botti}, \binits{G.}}, \betal:
\batitle{Bracs: A dataset for breast carcinoma subtyping in h\&e histology images}.
\bjtitle{Database}
\bvolume{2022},
\bfpage{093}
(\byear{2022})
\end{barticle}
\endbibitem

\bibitem[\protect\citeauthoryear{Aksac et~al.}{2019}]{aksac2019brecahad}
\begin{barticle}
\bauthor{\bsnm{Aksac}, \binits{A.}},
\bauthor{\bsnm{Demetrick}, \binits{D.J.}},
\bauthor{\bsnm{Ozyer}, \binits{T.}},
\bauthor{\bsnm{Alhajj}, \binits{R.}}:
\batitle{Brecahad: a dataset for breast cancer histopathological annotation and diagnosis}.
\bjtitle{BMC research notes}
\bvolume{12},
\bfpage{1}--\blpage{3}
(\byear{2019})
\end{barticle}
\endbibitem

\bibitem[\protect\citeauthoryear{Martel et~al.}{2019}]{martel2019assessment}
\begin{botherref}
\oauthor{\bsnm{Martel}, \binits{A.}},
\oauthor{\bsnm{Nofech-Mozes}, \binits{S.}},
\oauthor{\bsnm{Salama}, \binits{S.}},
\oauthor{\bsnm{Akbar}, \binits{S.}},
\oauthor{\bsnm{Peikari}, \binits{M.}}:
Assessment of residual breast cancer cellularity after neoadjuvant chemotherapy using digital pathology [data set].
The Cancer Imaging Archive
(2019)
\end{botherref}
\endbibitem

\bibitem[\protect\citeauthoryear{{National Cancer Institute Clinical Proteomic Tumor Analysis Consortium}}{2020}]{cptac2020}
\begin{botherref}
\oauthor{\bsnm{{National Cancer Institute Clinical Proteomic Tumor Analysis Consortium}}}:
The Clinical Proteomic Tumor Analysis Consortium Breast Invasive Carcinoma Collection (CPTAC-BRCA).
The Cancer Imaging Archive.
Accessed 07 Jul. 2023
(2020).
\url{https://wiki.cancerimagingarchive.net/pages/viewpage.action?pageId=70227748}
\end{botherref}
\endbibitem

\bibitem[\protect\citeauthoryear{Yan et~al.}{2021}]{yan2021richer}
\begin{barticle}
\bauthor{\bsnm{Yan}, \binits{R.}},
\bauthor{\bsnm{Zhang}, \binits{F.}},
\bauthor{\bsnm{Rao}, \binits{X.}},
\bauthor{\bsnm{Lv}, \binits{Z.}},
\bauthor{\bsnm{Li}, \binits{J.}},
\bauthor{\bsnm{Zhang}, \binits{L.}},
\bauthor{\bsnm{Liang}, \binits{S.}},
\bauthor{\bsnm{Li}, \binits{Y.}},
\bauthor{\bsnm{Ren}, \binits{F.}},
\bauthor{\bsnm{Zheng}, \binits{C.}}, \betal:
\batitle{Richer fusion network for breast cancer classification based on multimodal data}.
\bjtitle{BMC Medical Informatics and Decision Making}
\bvolume{21},
\bfpage{1}--\blpage{15}
(\byear{2021})
\end{barticle}
\endbibitem

\bibitem[\protect\citeauthoryear{{vEarly Breast Cancer Core-Needle Biopsy WSI (BCNB)}}{2022}]{bcnb2022}
\begin{botherref}
\oauthor{\bsnm{{vEarly Breast Cancer Core-Needle Biopsy WSI (BCNB)}}}:
Grand Challenge.
\url{https://bcnb.grand-challenge.org/}.
Accessed 07 Jul. 2023
(2022)
\end{botherref}
\endbibitem

\bibitem[\protect\citeauthoryear{Huang et~al.}{2023}]{huang2023artificial}
\begin{barticle}
\bauthor{\bsnm{Huang}, \binits{Z.}},
\bauthor{\bsnm{Shao}, \binits{W.}},
\bauthor{\bsnm{Han}, \binits{Z.}},
\bauthor{\bsnm{Alkashash}, \binits{A.M.}},
\bauthor{\bsnm{Sancha}, \binits{C.}},
\bauthor{\bsnm{Parwani}, \binits{A.V.}},
\bauthor{\bsnm{Nitta}, \binits{H.}},
\bauthor{\bsnm{Hou}, \binits{Y.}},
\bauthor{\bsnm{Wang}, \binits{T.}},
\bauthor{\bsnm{Salama}, \binits{P.}}, \betal:
\batitle{Artificial intelligence reveals features associated with breast cancer neoadjuvant chemotherapy responses from multi-stain histopathologic images}.
\bjtitle{NPJ Precision Oncology}
\bvolume{7}(\bissue{1}),
\bfpage{14}
(\byear{2023})
\end{barticle}
\endbibitem

\bibitem[\protect\citeauthoryear{{The Genotype-Tissue Expression (GTEx)}}{}]{gtex2023}
\begin{botherref}
\oauthor{\bsnm{{The Genotype-Tissue Expression (GTEx)}}}:
GTEx Portal.
\url{https://gtexportal.org/home/histologyPage}.
Accessed 07 Jul. 2023
\end{botherref}
\endbibitem

\bibitem[\protect\citeauthoryear{Lipkova et~al.}{2022}]{lipkova2022artificial}
\begin{barticle}
\bauthor{\bsnm{Lipkova}, \binits{J.}},
\bauthor{\bsnm{Chen}, \binits{R.J.}},
\bauthor{\bsnm{Chen}, \binits{B.}},
\bauthor{\bsnm{Lu}, \binits{M.Y.}},
\bauthor{\bsnm{Barbieri}, \binits{M.}},
\bauthor{\bsnm{Shao}, \binits{D.}},
\bauthor{\bsnm{Vaidya}, \binits{A.J.}},
\bauthor{\bsnm{Chen}, \binits{C.}},
\bauthor{\bsnm{Zhuang}, \binits{L.}},
\bauthor{\bsnm{Williamson}, \binits{D.F.}}, \betal:
\batitle{Artificial intelligence for multimodal data integration in oncology}.
\bjtitle{Cancer cell}
\bvolume{40}(\bissue{10}),
\bfpage{1095}--\blpage{1110}
(\byear{2022})
\end{barticle}
\endbibitem

\bibitem[\protect\citeauthoryear{Zhao et~al.}{2024}]{zhao2024deep}
\begin{botherref}
\oauthor{\bsnm{Zhao}, \binits{F.}},
\oauthor{\bsnm{Zhang}, \binits{C.}},
\oauthor{\bsnm{Geng}, \binits{B.}}:
Deep multimodal data fusion.
ACM Computing Surveys
(2024)
\end{botherref}
\endbibitem

\bibitem[\protect\citeauthoryear{Yellapragada et~al.}{2024}]{yellapragada2024pathldm}
\begin{bchapter}
\bauthor{\bsnm{Yellapragada}, \binits{S.}},
\bauthor{\bsnm{Graikos}, \binits{A.}},
\bauthor{\bsnm{Prasanna}, \binits{P.}},
\bauthor{\bsnm{Kurc}, \binits{T.}},
\bauthor{\bsnm{Saltz}, \binits{J.}},
\bauthor{\bsnm{Samaras}, \binits{D.}}:
\bctitle{Pathldm: Text conditioned latent diffusion model for histopathology}.
In: \bbtitle{Proceedings of the IEEE/CVF Winter Conference on Applications of Computer Vision},
pp. \bfpage{5182}--\blpage{5191}
(\byear{2024})
\end{bchapter}
\endbibitem

\bibitem[\protect\citeauthoryear{Ding et~al.}{2023}]{ding2023improving}
\begin{botherref}
\oauthor{\bsnm{Ding}, \binits{R.}},
\oauthor{\bsnm{Hall}, \binits{J.}},
\oauthor{\bsnm{Tenenholtz}, \binits{N.}},
\oauthor{\bsnm{Severson}, \binits{K.}}:
Improving mitosis detection on histopathology images using large vision-language models.
arXiv preprint arXiv:2310.07176
(2023)
\end{botherref}
\endbibitem

\bibitem[\protect\citeauthoryear{Lu et~al.}{2023}]{lu2023visual}
\begin{bchapter}
\bauthor{\bsnm{Lu}, \binits{M.Y.}},
\bauthor{\bsnm{Chen}, \binits{B.}},
\bauthor{\bsnm{Zhang}, \binits{A.}},
\bauthor{\bsnm{Williamson}, \binits{D.F.}},
\bauthor{\bsnm{Chen}, \binits{R.J.}},
\bauthor{\bsnm{Ding}, \binits{T.}},
\bauthor{\bsnm{Le}, \binits{L.P.}},
\bauthor{\bsnm{Chuang}, \binits{Y.-S.}},
\bauthor{\bsnm{Mahmood}, \binits{F.}}:
\bctitle{Visual language pretrained multiple instance zero-shot transfer for histopathology images}.
In: \bbtitle{Proceedings of the IEEE/CVF Conference on Computer Vision and Pattern Recognition},
pp. \bfpage{19764}--\blpage{19775}
(\byear{2023})
\end{bchapter}
\endbibitem

\bibitem[\protect\citeauthoryear{Arya and Saha}{2021}]{arya2021multi}
\begin{barticle}
\bauthor{\bsnm{Arya}, \binits{N.}},
\bauthor{\bsnm{Saha}, \binits{S.}}:
\batitle{Multi-modal advanced deep learning architectures for breast cancer survival prediction}.
\bjtitle{Knowledge-Based Systems}
\bvolume{221},
\bfpage{106965}
(\byear{2021})
\end{barticle}
\endbibitem

\bibitem[\protect\citeauthoryear{Sun et~al.}{2018}]{sun2018multimodal}
\begin{barticle}
\bauthor{\bsnm{Sun}, \binits{D.}},
\bauthor{\bsnm{Wang}, \binits{M.}},
\bauthor{\bsnm{Li}, \binits{A.}}:
\batitle{A multimodal deep neural network for human breast cancer prognosis prediction by integrating multi-dimensional data}.
\bjtitle{IEEE/ACM transactions on computational biology and bioinformatics}
\bvolume{16}(\bissue{3}),
\bfpage{841}--\blpage{850}
(\byear{2018})
\end{barticle}
\endbibitem

\bibitem[\protect\citeauthoryear{Tong et~al.}{2020}]{tong2020deep}
\begin{barticle}
\bauthor{\bsnm{Tong}, \binits{L.}},
\bauthor{\bsnm{Mitchel}, \binits{J.}},
\bauthor{\bsnm{Chatlin}, \binits{K.}},
\bauthor{\bsnm{Wang}, \binits{M.D.}}:
\batitle{Deep learning based feature-level integration of multi-omics data for breast cancer patients survival analysis}.
\bjtitle{BMC medical informatics and decision making}
\bvolume{20},
\bfpage{1}--\blpage{12}
(\byear{2020})
\end{barticle}
\endbibitem

\bibitem[\protect\citeauthoryear{Subramanian et~al.}{2021}]{subramanian2021multi}
\begin{botherref}
\oauthor{\bsnm{Subramanian}, \binits{V.}},
\oauthor{\bsnm{Syeda-Mahmood}, \binits{T.}},
\oauthor{\bsnm{Do}, \binits{M.N.}}:
Multi-modality fusion using canonical correlation analysis methods: Application in breast cancer survival prediction from histology and genomics.
arXiv preprint arXiv:2111.13987
(2021)
\end{botherref}
\endbibitem

\bibitem[\protect\citeauthoryear{Liu et~al.}{2022}]{liu2022hybrid}
\begin{barticle}
\bauthor{\bsnm{Liu}, \binits{T.}},
\bauthor{\bsnm{Huang}, \binits{J.}},
\bauthor{\bsnm{Liao}, \binits{T.}},
\bauthor{\bsnm{Pu}, \binits{R.}},
\bauthor{\bsnm{Liu}, \binits{S.}},
\bauthor{\bsnm{Peng}, \binits{Y.}}:
\batitle{A hybrid deep learning model for predicting molecular subtypes of human breast cancer using multimodal data}.
\bjtitle{Irbm}
\bvolume{43}(\bissue{1}),
\bfpage{62}--\blpage{74}
(\byear{2022})
\end{barticle}
\endbibitem

\bibitem[\protect\citeauthoryear{Howard et~al.}{2022}]{howard2022multimodal}
\begin{botherref}
\oauthor{\bsnm{Howard}, \binits{F.M.}},
\oauthor{\bsnm{Dolezal}, \binits{J.}},
\oauthor{\bsnm{Kochanny}, \binits{S.}},
\oauthor{\bsnm{Khramtsova}, \binits{G.}},
\oauthor{\bsnm{Vickery}, \binits{J.}},
\oauthor{\bsnm{Srisuwananukorn}, \binits{A.}},
\oauthor{\bsnm{Woodard}, \binits{A.}},
\oauthor{\bsnm{Chen}, \binits{N.}},
\oauthor{\bsnm{Nanda}, \binits{R.}},
\oauthor{\bsnm{Perou}, \binits{C.M.}}, et al.:
Multimodal prediction of breast cancer recurrence assays and risk of recurrence.
bioRxiv,
2022--07
(2022)
\end{botherref}
\endbibitem

\bibitem[\protect\citeauthoryear{Arya and Saha}{2021}]{arya2021generative}
\begin{barticle}
\bauthor{\bsnm{Arya}, \binits{N.}},
\bauthor{\bsnm{Saha}, \binits{S.}}:
\batitle{Generative incomplete multi-view prognosis predictor for breast cancer: Gimpp}.
\bjtitle{IEEE/ACM Transactions on Computational Biology and Bioinformatics}
\bvolume{19}(\bissue{4}),
\bfpage{2252}--\blpage{2263}
(\byear{2021})
\end{barticle}
\endbibitem

\bibitem[\protect\citeauthoryear{Arya and Saha}{2020}]{arya2020multi}
\begin{barticle}
\bauthor{\bsnm{Arya}, \binits{N.}},
\bauthor{\bsnm{Saha}, \binits{S.}}:
\batitle{Multi-modal classification for human breast cancer prognosis prediction: proposal of deep-learning based stacked ensemble model}.
\bjtitle{IEEE/ACM transactions on computational biology and bioinformatics}
\bvolume{19}(\bissue{2}),
\bfpage{1032}--\blpage{1041}
(\byear{2020})
\end{barticle}
\endbibitem

\bibitem[\protect\citeauthoryear{Furtney et~al.}{2023}]{furtney2023patient}
\begin{botherref}
\oauthor{\bsnm{Furtney}, \binits{I.}},
\oauthor{\bsnm{Bradley}, \binits{R.}},
\oauthor{\bsnm{Kabuka}, \binits{M.R.}}:
Patient graph deep learning to predict breast cancer molecular subtype.
IEEE/ACM transactions on computational biology and bioinformatics
(2023)
\end{botherref}
\endbibitem

\bibitem[\protect\citeauthoryear{Rani et~al.}{2023}]{rani2023diagnosis}
\begin{barticle}
\bauthor{\bsnm{Rani}, \binits{S.}},
\bauthor{\bsnm{Ahmad}, \binits{T.}},
\bauthor{\bsnm{Masood}, \binits{S.}},
\bauthor{\bsnm{Saxena}, \binits{C.}}:
\batitle{Diagnosis of breast cancer molecular subtypes using machine learning models on unimodal and multimodal datasets}.
\bjtitle{Neural Computing and Applications}
\bvolume{35}(\bissue{34}),
\bfpage{24109}--\blpage{24121}
(\byear{2023})
\end{barticle}
\endbibitem

\bibitem[\protect\citeauthoryear{Kayikci and Khoshgoftaar}{2023}]{kayikci2023breast}
\begin{barticle}
\bauthor{\bsnm{Kayikci}, \binits{S.}},
\bauthor{\bsnm{Khoshgoftaar}, \binits{T.M.}}:
\batitle{Breast cancer prediction using gated attentive multimodal deep learning}.
\bjtitle{Journal of Big Data}
\bvolume{10}(\bissue{1}),
\bfpage{62}
(\byear{2023})
\end{barticle}
\endbibitem

\bibitem[\protect\citeauthoryear{Arya et~al.}{2023}]{arya2023improving}
\begin{barticle}
\bauthor{\bsnm{Arya}, \binits{N.}},
\bauthor{\bsnm{Saha}, \binits{S.}},
\bauthor{\bsnm{Mathur}, \binits{A.}},
\bauthor{\bsnm{Saha}, \binits{S.}}:
\batitle{Improving the robustness and stability of a machine learning model for breast cancer prognosis through the use of multi-modal classifiers}.
\bjtitle{Scientific Reports}
\bvolume{13}(\bissue{1}),
\bfpage{4079}
(\byear{2023})
\end{barticle}
\endbibitem

\bibitem[\protect\citeauthoryear{Mondol et~al.}{2024}]{mondol2024mm}
\begin{botherref}
\oauthor{\bsnm{Mondol}, \binits{R.K.}},
\oauthor{\bsnm{Millar}, \binits{E.K.}},
\oauthor{\bsnm{Sowmya}, \binits{A.}},
\oauthor{\bsnm{Meijering}, \binits{E.}}:
Mm-survnet: Deep learning-based survival risk stratification in breast cancer through multimodal data fusion.
arXiv preprint arXiv:2402.11788
(2024)
\end{botherref}
\endbibitem

\bibitem[\protect\citeauthoryear{Huang et~al.}{2024}]{huang2024multimodal}
\begin{bchapter}
\bauthor{\bsnm{Huang}, \binits{S.}},
\bauthor{\bsnm{Liu}, \binits{Z.}},
\bauthor{\bsnm{Liu}, \binits{Z.}}:
\bctitle{Multimodal siamese model for breast cancer survival prediction}.
In: \bbtitle{2024 4th International Conference on Neural Networks, Information and Communication (NNICE)},
pp. \bfpage{925}--\blpage{928}
(\byear{2024}).
\bcomment{IEEE}
\end{bchapter}
\endbibitem

\bibitem[\protect\citeauthoryear{Li and Nabavi}{2024}]{li2024multimodal}
\begin{barticle}
\bauthor{\bsnm{Li}, \binits{B.}},
\bauthor{\bsnm{Nabavi}, \binits{S.}}:
\batitle{A multimodal graph neural network framework for cancer molecular subtype classification}.
\bjtitle{BMC bioinformatics}
\bvolume{25}(\bissue{1}),
\bfpage{27}
(\byear{2024})
\end{barticle}
\endbibitem

\bibitem[\protect\citeauthoryear{Gu et~al.}{2020}]{gu2020case}
\begin{barticle}
\bauthor{\bsnm{Gu}, \binits{D.}},
\bauthor{\bsnm{Su}, \binits{K.}},
\bauthor{\bsnm{Zhao}, \binits{H.}}:
\batitle{A case-based ensemble learning system for explainable breast cancer recurrence prediction}.
\bjtitle{Artificial Intelligence in Medicine}
\bvolume{107},
\bfpage{101858}
(\byear{2020})
\end{barticle}
\endbibitem

\bibitem[\protect\citeauthoryear{Kabak{\c{c}}{\i} et~al.}{2021}]{kabakcci2021automated}
\begin{barticle}
\bauthor{\bsnm{Kabak{\c{c}}{\i}}, \binits{K.A.}},
\bauthor{\bsnm{{\c{C}}ak{\i}r}, \binits{A.}},
\bauthor{\bsnm{T{\"u}rkmen}, \binits{{\. I}.}},
\bauthor{\bsnm{T{\"o}reyin}, \binits{B.U.}},
\bauthor{\bsnm{{\c{C}}apar}, \binits{A.}}:
\batitle{Automated scoring of cerbb2/her2 receptors using histogram based analysis of immunohistochemistry breast cancer tissue images}.
\bjtitle{Biomedical Signal Processing and Control}
\bvolume{69},
\bfpage{102924}
(\byear{2021})
\end{barticle}
\endbibitem

\bibitem[\protect\citeauthoryear{Maleki et~al.}{2023}]{maleki2023breast}
\begin{barticle}
\bauthor{\bsnm{Maleki}, \binits{A.}},
\bauthor{\bsnm{Raahemi}, \binits{M.}},
\bauthor{\bsnm{Nasiri}, \binits{H.}}:
\batitle{Breast cancer diagnosis from histopathology images using deep neural network and xgboost}.
\bjtitle{Biomedical Signal Processing and Control}
\bvolume{86},
\bfpage{105152}
(\byear{2023})
\end{barticle}
\endbibitem

\bibitem[\protect\citeauthoryear{Peta and Koppu}{2024}]{peta2024explainable}
\begin{barticle}
\bauthor{\bsnm{Peta}, \binits{J.}},
\bauthor{\bsnm{Koppu}, \binits{S.}}:
\batitle{Explainable soft attentive efficientnet for breast cancer classification in histopathological images}.
\bjtitle{Biomedical Signal Processing and Control}
\bvolume{90},
\bfpage{105828}
(\byear{2024})
\end{barticle}
\endbibitem

\bibitem[\protect\citeauthoryear{Jaume et~al.}{2020}]{jaume2020towards}
\begin{botherref}
\oauthor{\bsnm{Jaume}, \binits{G.}},
\oauthor{\bsnm{Pati}, \binits{P.}},
\oauthor{\bsnm{Foncubierta-Rodriguez}, \binits{A.}},
\oauthor{\bsnm{Feroce}, \binits{F.}},
\oauthor{\bsnm{Scognamiglio}, \binits{G.}},
\oauthor{\bsnm{Anniciello}, \binits{A.M.}},
\oauthor{\bsnm{Thiran}, \binits{J.-P.}},
\oauthor{\bsnm{Goksel}, \binits{O.}},
\oauthor{\bsnm{Gabrani}, \binits{M.}}:
Towards explainable graph representations in digital pathology.
arXiv preprint arXiv:2007.00311
(2020)
\end{botherref}
\endbibitem

\bibitem[\protect\citeauthoryear{Maouche et~al.}{2023}]{maouche2023explainable}
\begin{botherref}
\oauthor{\bsnm{Maouche}, \binits{I.}},
\oauthor{\bsnm{Terrissa}, \binits{L.S.}},
\oauthor{\bsnm{Benmohammed}, \binits{K.}},
\oauthor{\bsnm{Zerhouni}, \binits{N.}}:
An explainable ai approach for breast cancer metastasis prediction based on clinicopathological data.
IEEE Transactions on Biomedical Engineering
(2023)
\end{botherref}
\endbibitem

\bibitem[\protect\citeauthoryear{Altini et~al.}{2023}]{altini2023tumor}
\begin{barticle}
\bauthor{\bsnm{Altini}, \binits{N.}},
\bauthor{\bsnm{Puro}, \binits{E.}},
\bauthor{\bsnm{Taccogna}, \binits{M.G.}},
\bauthor{\bsnm{Marino}, \binits{F.}},
\bauthor{\bsnm{De~Summa}, \binits{S.}},
\bauthor{\bsnm{Saponaro}, \binits{C.}},
\bauthor{\bsnm{Mattioli}, \binits{E.}},
\bauthor{\bsnm{Zito}, \binits{F.A.}},
\bauthor{\bsnm{Bevilacqua}, \binits{V.}}:
\batitle{Tumor cellularity assessment of breast histopathological slides via instance segmentation and pathomic features explainability}.
\bjtitle{Bioengineering}
\bvolume{10}(\bissue{4}),
\bfpage{396}
(\byear{2023})
\end{barticle}
\endbibitem

\bibitem[\protect\citeauthoryear{Liu and Hu}{2022}]{liu2022extendable}
\begin{barticle}
\bauthor{\bsnm{Liu}, \binits{Q.}},
\bauthor{\bsnm{Hu}, \binits{P.}}:
\batitle{Extendable and explainable deep learning for pan-cancer radiogenomics research}.
\bjtitle{Current opinion in chemical biology}
\bvolume{66},
\bfpage{102111}
(\byear{2022})
\end{barticle}
\endbibitem

\bibitem[\protect\citeauthoryear{Holzinger et~al.}{2021}]{holzinger2021towards}
\begin{barticle}
\bauthor{\bsnm{Holzinger}, \binits{A.}},
\bauthor{\bsnm{Malle}, \binits{B.}},
\bauthor{\bsnm{Saranti}, \binits{A.}},
\bauthor{\bsnm{Pfeifer}, \binits{B.}}:
\batitle{Towards multi-modal causability with graph neural networks enabling information fusion for explainable ai}.
\bjtitle{Information Fusion}
\bvolume{71},
\bfpage{28}--\blpage{37}
(\byear{2021})
\end{barticle}
\endbibitem

\bibitem[\protect\citeauthoryear{Zhang et~al.}{2021}]{zhang2021dmrfnet}
\begin{barticle}
\bauthor{\bsnm{Zhang}, \binits{W.}},
\bauthor{\bsnm{Yu}, \binits{J.}},
\bauthor{\bsnm{Zhao}, \binits{W.}},
\bauthor{\bsnm{Ran}, \binits{C.}}:
\batitle{Dmrfnet: deep multimodal reasoning and fusion for visual question answering and explanation generation}.
\bjtitle{Information Fusion}
\bvolume{72},
\bfpage{70}--\blpage{79}
(\byear{2021})
\end{barticle}
\endbibitem

\bibitem[\protect\citeauthoryear{Kang et~al.}{2023}]{kang2023learning}
\begin{barticle}
\bauthor{\bsnm{Kang}, \binits{S.}},
\bauthor{\bsnm{Chen}, \binits{Z.}},
\bauthor{\bsnm{Li}, \binits{L.}},
\bauthor{\bsnm{Lu}, \binits{W.}},
\bauthor{\bsnm{Qi}, \binits{X.S.}},
\bauthor{\bsnm{Tan}, \binits{S.}}:
\batitle{Learning feature fusion via an interpretation method for tumor segmentation on pet/ct}.
\bjtitle{Applied Soft Computing}
\bvolume{148},
\bfpage{110825}
(\byear{2023})
\end{barticle}
\endbibitem

\bibitem[\protect\citeauthoryear{Krishna et~al.}{2023}]{krishna2023interpretable}
\begin{barticle}
\bauthor{\bsnm{Krishna}, \binits{S.}},
\bauthor{\bsnm{Suganthi}, \binits{S.}},
\bauthor{\bsnm{Bhavsar}, \binits{A.}},
\bauthor{\bsnm{Yesodharan}, \binits{J.}},
\bauthor{\bsnm{Krishnamoorthy}, \binits{S.}}:
\batitle{An interpretable decision-support model for breast cancer diagnosis using histopathology images}.
\bjtitle{Journal of Pathology Informatics}
\bvolume{14},
\bfpage{100319}
(\byear{2023})
\end{barticle}
\endbibitem

\bibitem[\protect\citeauthoryear{Guo et~al.}{2024}]{guo2024histgen}
\begin{botherref}
\oauthor{\bsnm{Guo}, \binits{Z.}},
\oauthor{\bsnm{Ma}, \binits{J.}},
\oauthor{\bsnm{Xu}, \binits{Y.}},
\oauthor{\bsnm{Wang}, \binits{Y.}},
\oauthor{\bsnm{Wang}, \binits{L.}},
\oauthor{\bsnm{Chen}, \binits{H.}}:
Histgen: Histopathology report generation via local-global feature encoding and cross-modal context interaction.
arXiv preprint arXiv:2403.05396
(2024)
\end{botherref}
\endbibitem

\bibitem[\protect\citeauthoryear{Hartsock and Rasool}{2024}]{hartsock2024vision}
\begin{botherref}
\oauthor{\bsnm{Hartsock}, \binits{I.}},
\oauthor{\bsnm{Rasool}, \binits{G.}}:
Vision-language models for medical report generation and visual question answering: A review.
arXiv preprint arXiv:2403.02469
(2024)
\end{botherref}
\endbibitem

\bibitem[\protect\citeauthoryear{Hu et~al.}{2024}]{hu2024histopathology}
\begin{barticle}
\bauthor{\bsnm{Hu}, \binits{D.}},
\bauthor{\bsnm{Jiang}, \binits{Z.}},
\bauthor{\bsnm{Shi}, \binits{J.}},
\bauthor{\bsnm{Xie}, \binits{F.}},
\bauthor{\bsnm{Wu}, \binits{K.}},
\bauthor{\bsnm{Tang}, \binits{K.}},
\bauthor{\bsnm{Cao}, \binits{M.}},
\bauthor{\bsnm{Huai}, \binits{J.}},
\bauthor{\bsnm{Zheng}, \binits{Y.}}:
\batitle{Histopathology language-image representation learning for fine-grained digital pathology cross-modal retrieval}.
\bjtitle{Medical Image Analysis}
\bvolume{95},
\bfpage{103163}
(\byear{2024})
\end{barticle}
\endbibitem

\bibitem[\protect\citeauthoryear{Van~Rijthoven et~al.}{2021}]{van2021hooknet}
\begin{barticle}
\bauthor{\bsnm{Van~Rijthoven}, \binits{M.}},
\bauthor{\bsnm{Balkenhol}, \binits{M.}},
\bauthor{\bsnm{Sili{\c{n}}a}, \binits{K.}},
\bauthor{\bsnm{Van Der~Laak}, \binits{J.}},
\bauthor{\bsnm{Ciompi}, \binits{F.}}:
\batitle{Hooknet: Multi-resolution convolutional neural networks for semantic segmentation in histopathology whole-slide images}.
\bjtitle{Medical image analysis}
\bvolume{68},
\bfpage{101890}
(\byear{2021})
\end{barticle}
\endbibitem

\bibitem[\protect\citeauthoryear{Held et~al.}{2024}]{held2024x}
\begin{botherref}
\oauthor{\bsnm{Held}, \binits{J.}},
\oauthor{\bsnm{Itani}, \binits{H.}},
\oauthor{\bsnm{Cioppa}, \binits{A.}},
\oauthor{\bsnm{Giancola}, \binits{S.}},
\oauthor{\bsnm{Ghanem}, \binits{B.}},
\oauthor{\bsnm{Van~Droogenbroeck}, \binits{M.}}:
X-vars: Introducing explainability in football refereeing with multi-modal large language model.
arXiv preprint arXiv:2404.06332
(2024)
\end{botherref}
\endbibitem

\bibitem[\protect\citeauthoryear{Bousselham et~al.}{2024}]{bousselham2024legrad}
\begin{botherref}
\oauthor{\bsnm{Bousselham}, \binits{W.}},
\oauthor{\bsnm{Boggust}, \binits{A.}},
\oauthor{\bsnm{Chaybouti}, \binits{S.}},
\oauthor{\bsnm{Strobelt}, \binits{H.}},
\oauthor{\bsnm{Kuehne}, \binits{H.}}:
Legrad: An explainability method for vision transformers via feature formation sensitivity.
arXiv preprint arXiv:2404.03214
(2024)
\end{botherref}
\endbibitem

\bibitem[\protect\citeauthoryear{Vaswani et~al.}{2017}]{vaswani2017attention}
\begin{botherref}
\oauthor{\bsnm{Vaswani}, \binits{A.}},
\oauthor{\bsnm{Shazeer}, \binits{N.}},
\oauthor{\bsnm{Parmar}, \binits{N.}},
\oauthor{\bsnm{Uszkoreit}, \binits{J.}},
\oauthor{\bsnm{Jones}, \binits{L.}},
\oauthor{\bsnm{Gomez}, \binits{A.N.}},
\oauthor{\bsnm{Kaiser}, \binits{{\L}.}},
\oauthor{\bsnm{Polosukhin}, \binits{I.}}:
Attention is all you need.
Advances in neural information processing systems
\textbf{30}
(2017)
\end{botherref}
\endbibitem

\bibitem[\protect\citeauthoryear{Selvaraju et~al.}{2017}]{selvaraju2017grad}
\begin{bchapter}
\bauthor{\bsnm{Selvaraju}, \binits{R.R.}},
\bauthor{\bsnm{Cogswell}, \binits{M.}},
\bauthor{\bsnm{Das}, \binits{A.}},
\bauthor{\bsnm{Vedantam}, \binits{R.}},
\bauthor{\bsnm{Parikh}, \binits{D.}},
\bauthor{\bsnm{Batra}, \binits{D.}}:
\bctitle{Grad-cam: Visual explanations from deep networks via gradient-based localization}.
In: \bbtitle{Proceedings of the IEEE International Conference on Computer Vision},
pp. \bfpage{618}--\blpage{626}
(\byear{2017})
\end{bchapter}
\endbibitem

\bibitem[\protect\citeauthoryear{Chen et~al.}{2024}]{chen2024actnet}
\begin{barticle}
\bauthor{\bsnm{Chen}, \binits{H.}},
\bauthor{\bsnm{Zhang}, \binits{X.}},
\bauthor{\bsnm{Guo}, \binits{Z.}},
\bauthor{\bsnm{Ying}, \binits{N.}},
\bauthor{\bsnm{Yang}, \binits{M.}},
\bauthor{\bsnm{Guo}, \binits{C.}}:
\batitle{Actnet: Attention based cnn and transformer network for respiratory rate estimation}.
\bjtitle{Biomedical Signal Processing and Control}
\bvolume{96},
\bfpage{106497}
(\byear{2024})
\end{barticle}
\endbibitem

\bibitem[\protect\citeauthoryear{Tizhoosh and Pantanowitz}{2024}]{tizhoosh2024image}
\begin{botherref}
\oauthor{\bsnm{Tizhoosh}, \binits{H.}},
\oauthor{\bsnm{Pantanowitz}, \binits{L.}}:
On image search in histopathology.
Journal of Pathology Informatics,
100375
(2024)
\end{botherref}
\endbibitem

\bibitem[\protect\citeauthoryear{Werner et~al.}{2022}]{werner2022ability}
\begin{botherref}
\oauthor{\bsnm{Werner}, \binits{P.}},
\oauthor{\bsnm{Zapaishchykova}, \binits{A.}},
\oauthor{\bsnm{Ratan}, \binits{U.}}:
The ability of image-language explainable models to resemble domain expertise.
arXiv preprint arXiv:2209.09310
(2022)
\end{botherref}
\endbibitem

\bibitem[\protect\citeauthoryear{Nguyen et~al.}{2024}]{nguyen2024langxai}
\begin{botherref}
\oauthor{\bsnm{Nguyen}, \binits{T.T.H.}},
\oauthor{\bsnm{Clement}, \binits{T.}},
\oauthor{\bsnm{Nguyen}, \binits{P.T.L.}},
\oauthor{\bsnm{Kemmerzell}, \binits{N.}},
\oauthor{\bsnm{Truong}, \binits{V.B.}},
\oauthor{\bsnm{Nguyen}, \binits{V.T.K.}},
\oauthor{\bsnm{Abdelaal}, \binits{M.}},
\oauthor{\bsnm{Cao}, \binits{H.}}:
Langxai: Integrating large vision models for generating textual explanations to enhance explainability in visual perception tasks.
arXiv preprint arXiv:2402.12525
(2024)
\end{botherref}
\endbibitem

\bibitem[\protect\citeauthoryear{Rehman~Hashmi et~al.}{2024}]{rehman2024envisioning}
\begin{botherref}
\oauthor{\bsnm{Rehman~Hashmi}, \binits{A.U.}},
\oauthor{\bsnm{Mahapatra}, \binits{D.}},
\oauthor{\bsnm{Yaqub}, \binits{M.}}:
Envisioning medclip: A deep dive into explainability for medical vision-language models.
arXiv e-prints,
2403
(2024)
\end{botherref}
\endbibitem

\bibitem[\protect\citeauthoryear{Ben Melech~Stan et~al.}{2024}]{ben2024lvlm}
\begin{botherref}
\oauthor{\bsnm{Ben Melech~Stan}, \binits{G.}},
\oauthor{\bsnm{Yehezkel~Rohekar}, \binits{R.}},
\oauthor{\bsnm{Gurwicz}, \binits{Y.}},
\oauthor{\bsnm{Olson}, \binits{M.L.}},
\oauthor{\bsnm{Bhiwandiwalla}, \binits{A.}},
\oauthor{\bsnm{Aflalo}, \binits{E.}},
\oauthor{\bsnm{Wu}, \binits{C.}},
\oauthor{\bsnm{Duan}, \binits{N.}},
\oauthor{\bsnm{Tseng}, \binits{S.-Y.}},
\oauthor{\bsnm{Lal}, \binits{V.}}:
Lvlm-intrepret: An interpretability tool for large vision-language models.
arXiv e-prints,
2404
(2024)
\end{botherref}
\endbibitem

\bibitem[\protect\citeauthoryear{Yang et~al.}{2023}]{yang2023neural}
\begin{botherref}
\oauthor{\bsnm{Yang}, \binits{X.}},
\oauthor{\bsnm{Liu}, \binits{F.}},
\oauthor{\bsnm{Lin}, \binits{G.}}:
Neural logic vision language explainer.
IEEE Transactions on Multimedia
(2023)
\end{botherref}
\endbibitem

\bibitem[\protect\citeauthoryear{Sammani and Deligiannis}{2023}]{sammani2023uni}
\begin{bchapter}
\bauthor{\bsnm{Sammani}, \binits{F.}},
\bauthor{\bsnm{Deligiannis}, \binits{N.}}:
\bctitle{Uni-nlx: Unifying textual explanations for vision and vision-language tasks}.
In: \bbtitle{Proceedings of the IEEE/CVF International Conference on Computer Vision},
pp. \bfpage{4634}--\blpage{4639}
(\byear{2023})
\end{bchapter}
\endbibitem

\bibitem[\protect\citeauthoryear{Biswal et~al.}{2020}]{biswal2020clara}
\begin{bchapter}
\bauthor{\bsnm{Biswal}, \binits{S.}},
\bauthor{\bsnm{Xiao}, \binits{C.}},
\bauthor{\bsnm{Glass}, \binits{L.M.}},
\bauthor{\bsnm{Westover}, \binits{B.}},
\bauthor{\bsnm{Sun}, \binits{J.}}:
\bctitle{Clara: clinical report auto-completion}.
In: \bbtitle{Proceedings of The Web Conference 2020},
pp. \bfpage{541}--\blpage{550}
(\byear{2020})
\end{bchapter}
\endbibitem

\end{thebibliography}

\end{document}